\documentclass[12pt,letterpaper]{article}
\usepackage[T1]{fontenc}
\usepackage[utf8]{inputenc}
\usepackage[a4paper, total={7in, 10in}]{geometry}

\usepackage{graphicx}
\usepackage{helvet}
\usepackage{multirow}
\usepackage{booktabs}
\usepackage{float}
\usepackage{authblk}
\usepackage{hyperref}
\usepackage{amsmath}
\usepackage{amssymb}
\usepackage[super,comma,sort&compress]{natbib}
\makeatletter
\renewcommand{\maketitle}{\bgroup\setlength{\parindent}{0pt}
\begin{flushleft}
  \textbf{\@title}

  \@author
\end{flushleft}\egroup}
\makeatother

\title{From Small to Large: Assessing Medical Reasoning, Benchmark Contamination, and Model Size in Language Models}
\date{}

\author[1]{Navapat Nananukul}
\author[1,2,*]{Mayank Kejriwal}

\affil[1]{Information Sciences Institute, University of Southern California, Marina del Rey, CA, USA}
\affil[2]{Lead contact}

\affil[*]{Correspondence: kejriwal@isi.edu, nananuku@isi.edu}

\begin{document}

\maketitle

\section*{SUMMARY}

Large language models (LLMs) now answer medical exam questions at or above expert level, but it is unclear whether such scores reflect clinical competence, or are instead affected by the format of the questions being asked (e.g., multiple-choice versus open-ended), prior exposure to the benchmarks, and the difficulty of the questions. These benchmarks are also public and widely used online, which may be resulting in LLM memorization. Models may perform better on questions that they have seen before. It is also not evident whether smaller models, which are more locally deployable in HIPAA-compliant hospital settings, have achieved human performance. In this study, we evaluate fourteen current models from six leading providers, one smaller and one larger model from each provider with two additional open-weight models, on 3,076 questions from four medical benchmarks (MedQA-USMLE, MedMCQA, PubMedQA, and MedBullets) under standard multiple choice, the same questions with the answer choices withheld, and reconstruction of a hidden question from its choices alone. In the standard multiple-choice setting, the strongest models exceed the human-expert baseline on MedQA-USMLE and PubMedQA, but the smaller models trail their larger models by 8.7 percentage points and only two of the six pass the human-expert bar on the licensing examination. Withholding the answer choices, however, makes every model struggle: mean accuracy falls from 82.3\% to 50.9\%, and the two largest drops belong to the medically specialised models. On memorization, the strongest models reconstruct one question in four from the answer choices alone, yet accuracy holds on the questions they cannot reconstruct. We further weigh accuracy against inference cost and test whether ensembling several models improves medical question answering.

\section*{KEYWORDS}

large language models; medical question answering; benchmark evaluation; reanalysis; question answering; open-ended evaluation; data contamination; memorization; LLM-as-a-judge

\section*{INTRODUCTION}

Large language models (LLMs) have become common tools for natural language generation and question answering. Scaling Transformer-based models to hundreds of billions of parameters and training them on internet-scale text has produced sustained gains \cite{vaswani2017attention,peters-etal-2018-deep,devlin2019bert,brown2020language}. A defining recent capability is explicit reasoning, and it has arrived through a number of advances such as Chain-of-thought prompting, \cite{wei2022chain}, Self-consistency \cite{wang2022self}, Tree-of-Thought \cite{yao2023tree}, and larger context windows \cite{an2024make}. Current models now generate step-by-step reasoning without being prompted for it, because it has been trained into them directly \cite{wang2024chain,chen2025towards}. Because of these advances, and because models have also grown larger, performance has improved for both small and large models: the smaller models available today reach the scores that large models reached a few years ago, while the large models of today still generally outperform the small ones.

Despite this rapid progress, medicine remains a challenging domain, as answering a clinical question demands specialized knowledge, careful reading of the case, and multi-step reasoning \cite{lucas2024reasoning,lievin2024can,singhal2025toward}. Specialized benchmarks were built to test these abilities together, and three have come to define the task: MedQA-USMLE \cite{app11146421}, drawn from the United States Medical Licensing Examination; MedMCQA \cite{pal2022medmcqa}, from Indian medical entrance exams; and PubMedQA \cite{jin2019pubmedqa}, which asks for a \emph{yes}, \emph{no}, or \emph{maybe} judgment about a biomedical abstract. A more recent set, MedBullets\cite{chen2024medbullets}, poses clinical cases written for USMLE board preparation.

A high score on these benchmarks, however, does not tell the whole story. The standard question-answering setup records only whether a model selected the correct choice, not how it got there. A model may reason through the case, or it may recognize the answer from the choices placed in front of it, or it may simply remember a question it encountered during training: these benchmarks are public and widely mirrored online, and models are known to reproduce text they were trained on \cite{ishihara-2023-training,golchin2024time}. All three paths yield the same score while implying very different clinical ability. Understanding what such a score actually measures therefore requires looking past the standard setup and testing models under conditions that separate reasoning from memorization \cite{zhang2024can,xie2024memorization}.

While benchmark scores indicate which models perform well on clinical question answering, not every model can be used in practice, for compliance reasons \cite{marks2023ai,neupane2025towards}. Protected health information generally cannot leave an institution's control, and as a consequence, a system that runs on-premises inside a HIPAA-compliant boundary is often the only kind a clinical team can use \cite{tripathi2026hipaa}. Small, open-weight language models are what make that possible \cite{gondara2025small,garg2025rise}: they fit on hardware a hospital already owns, cost far less per query at the volume a busy service generates, and can be served without a commercial API contract or an outbound network path for patient data. Those advantages compound where budgets are smallest---rural and community hospitals, which have the least access to specialist consultation and the most to gain from decision support. Whether a 9B model is close enough to a frontier model is therefore not an academic question. It is also possible that the distance between the two only looks small because of the way current benchmarks measure it.


In this study, we evaluate fourteen closed- and open-weight models from six providers on 3,076 questions from four benchmarks: MedQA-USMLE, MedMCQA, PubMedQA, and MedBullets. For each provider, we select a smaller and a larger model, where size is defined relative to the other model in the pair. For the four pairs with publicly reported parameter counts (DeepSeek, Meta, Mistral, and Google), the larger model has both more active and total parameters. For OpenAI and Qwen, the larger model's parameter count is undisclosed, so we assign it based on the provider's positioning of the model (a lightweight model and a flagship model). Because these two larger models are closed-weight, we also ran a third, open-weight model of reported size for each of the two providers to study the effect of model size with models whose size is known. 

We address four research questions: how far current reasoning models, both large and small, have improved over the previous generation (RQ1); whether that performance holds when the answer choices are withheld (RQ2); whether LLMs can reconstruct a question when shown only its answer choices (RQ3); and how accuracy differs between the questions a model can and cannot reconstruct (RQ4). To address these questions, we evaluate each model under three settings. The first is the standard multiple-choice format, following prior work \cite{lievin2024can}. The second is a generative setting, in which the answer choices are withheld and the free-text answer is evaluated. The third is a reconstruction setting, in which the model sees only the choices and must reconstruct the question. Prior work has either asked models open-ended questions\cite{johri2025evaluation,schmidgall2024agentclinic} or evaluated whether models have memorized benchmark questions\cite{golchin2024time,deng2024investigating}, typically examining only one aspect of performance and not always in the medical domain. Our analysis is novel in applying all three settings to the same questions and comprehensive sets of language models, revealing how much of a model's score depends on its knowledge, the answer choices, or familiarity with the question.

Our contributions are as follows.

\begin{itemize}
   \item We evaluate six smaller and six larger language models on four medical benchmarks. Each is evaluated with its inherent reasoning capability and with chain-of-thought reasoning.

   \item We propose a comprehensive protocol for evaluating clinical question answering, and assess every model under three question-answering settings: multiple-choice answering, generative answering, and reconstruction of a hidden question.

   \item We introduce question reconstruction as a proxy for benchmark memorization: the model is shown only the answer choices and attempts to regenerate the original question.

    \item We conduct rigorous analyses, including overall question-answering performance on three settings, a paired comparison of the smaller and larger model from the same provider, an analysis of model memorization and its relation to performance, a cost and performance analysis of LLMs, and the effect of LLM ensembling on clinical question answering. Throughout we report 95\% confidence intervals and test differences with paired $t$-tests, and $\chi^2$ tests.

   \item We release all artifacts underlying the experiments including the code that runs the medical question answering, the code that runs the judge together with its verdicts, the prompt templates for all three settings, and both grading rubrics. The evaluation and the grading material can each therefore be reproduced rather than reimplemented from description.
\end{itemize}

\section*{METHODS}

\begin{figure}[H]
\centering
\includegraphics[width=0.95\linewidth]{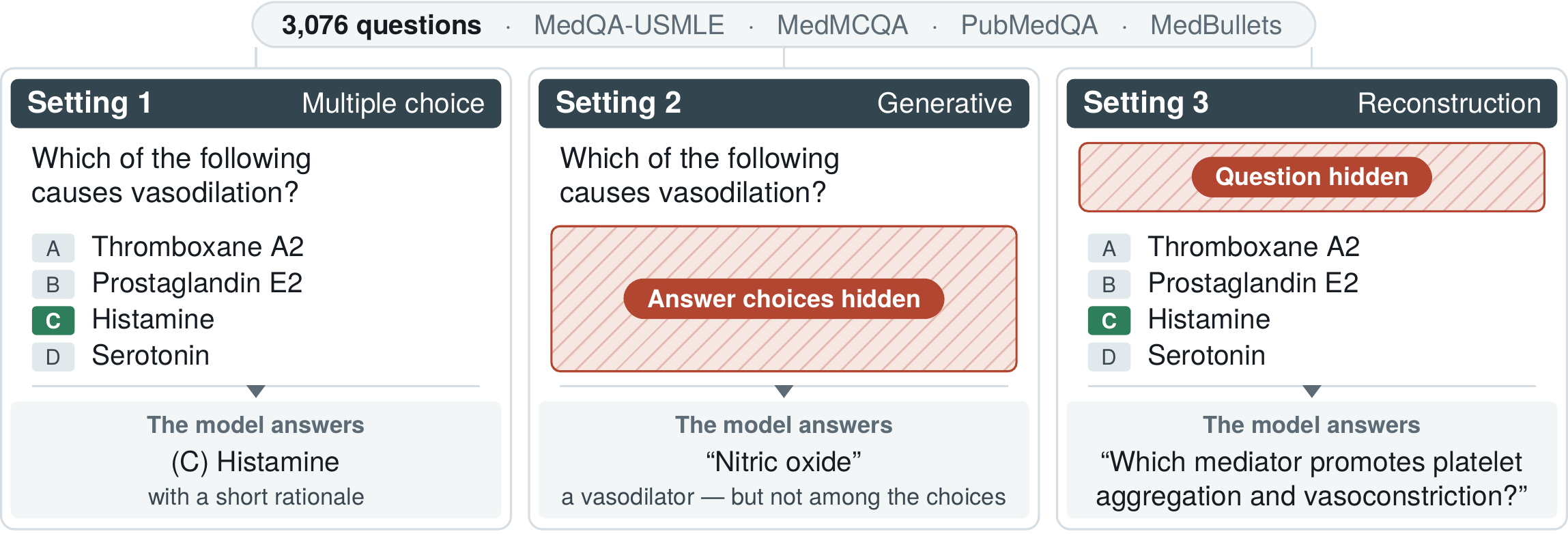}
\caption{The three evaluation settings, illustrated on one question from MedMCQA. Hatched panels mark the information withheld from the model in each setting.}
\label{fig:exp-design}
\end{figure}

We evaluate large language models (LLMs) across three experimental settings (Figure \ref{fig:exp-design}) designed to address our research questions. The first setting measures baseline performance (RQ1); the second investigates the extent to which performance depends on exposure to answer choices (RQ2); and the third examines potential contamination from benchmark familiarization (RQ3 and RQ4). This section describes our benchmarks and models, defines each experimental setting, details the evaluation protocols for open-ended outputs, and outlines the statistical analyses performed.

\subsection*{Datasets}

\begin{figure}[H]
\centering
\includegraphics[width=0.98\linewidth]{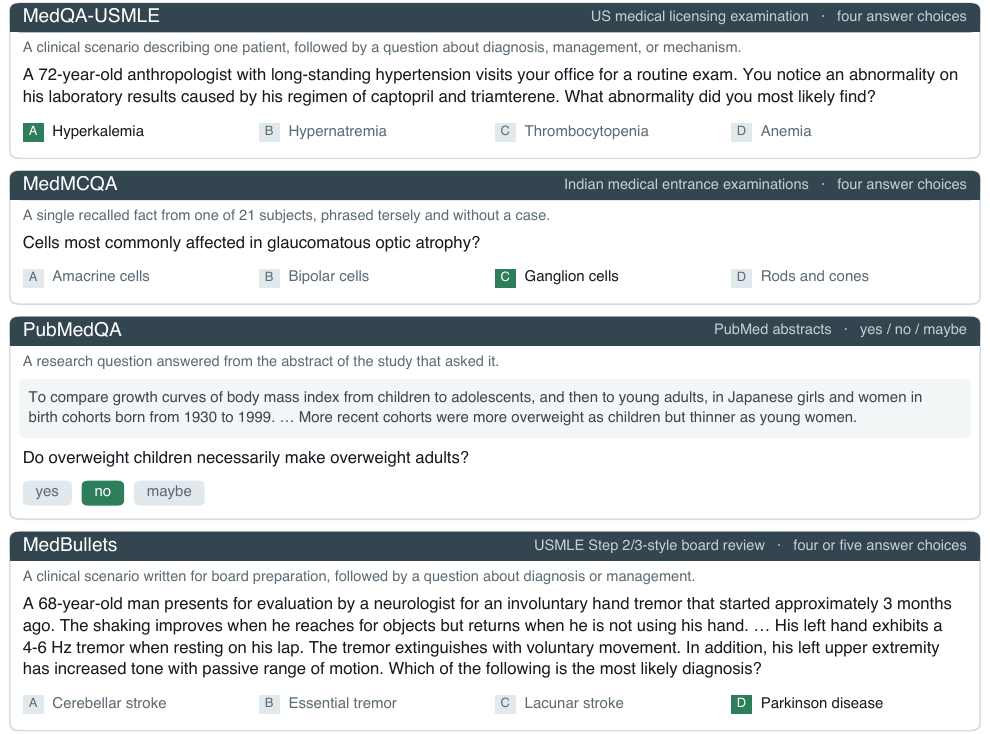}
\caption{One question example from each benchmark. Green marks the recorded answer key.}
\label{fig:bench-examples}
\end{figure}

We used all questions that are publicly available on the four benchmarks: MedQA-USMLE, MedMCQA, PubMedQA, and MedBullets (Table~\ref{tab:datasets}). As shown in Figure \ref{fig:bench-examples}, MedQA-USMLE and MedMCQA are four-choice multiple-choice sets from medical licensing and entrance exams. PubMedQA asks for a \emph{yes}, \emph{no}, or \emph{maybe} judgment about a biomedical abstract. MedBullets poses USMLE Step 2 and 3 style clinical cases written for board preparation, released in a four-choice and a five-choice version of the same questions. We obtained MedQA-USMLE and MedMCQA from Hugging Face and PubMedQA from the archive. For MedQA-USMLE this is the full 1,273-question test split. For PubMedQA it is the full 500-question expert-annotated test split. MedMCQA is the exception: its test-set labels are not public, because the official test set is scored only through leaderboard submission. We therefore evaluated MedMCQA on its validation split similar to the authors of the earlier evaluation \cite{lievin2024can}, and reused the same 1,000-question sample for comparability. MedBullets \cite{chen2024medbullets} contains 308 questions posted publicly since April 2022. In total, the accuracies are calculated by weighting over the 3,076 medical questions, and the analyses that withhold the choices or reconstruct the question use the 2,576 questions of MedQA-USMLE, MedMCQA and MedBullets.

\begin{table}[H]
\centering
\small
\caption{The four benchmarks and the evaluation sets used in this paper.}
\label{tab:datasets}
\begin{tabular}{p{0.13\linewidth}p{0.175\linewidth}p{0.175\linewidth}p{0.175\linewidth}p{0.175\linewidth}}
\toprule
 & \textbf{MedQA-USMLE}\cite{app11146421}
 & \textbf{MedMCQA}\cite{pal2022medmcqa} & \textbf{PubMedQA}\cite{jin2019pubmedqa} & \textbf{MedBullets}\cite{chen2024medbullets} \\
\midrule
\raggedright What the question asks & \raggedright A clinical scenario from one patient, followed by a question about diagnosis or health management & \raggedright A single fact drawn from one of 21 preclinical and clinical subjects & \raggedright Whether the abstract of a published study supports the research question that study asked & \raggedright A clinical scenario written for board preparation, followed by a question about diagnosis or management \tabularnewline
\addlinespace[2pt]
\raggedright Answer choices & \raggedright A/B/C/D & \raggedright A/B/C/D & \raggedright \emph{yes} / \emph{no} / \emph{maybe} & \raggedright A/B/C/D\tabularnewline
\raggedright Question source & \raggedright US licensing exam & \raggedright AIIMS/NEET PG & \raggedright PubMed abstracts & \raggedright MedBullets question bank \tabularnewline
\addlinespace[6pt]
\raggedright No. of questions & \raggedright 1,273 & \raggedright 1,000 & \raggedright 500 & \raggedright 303 \tabularnewline
\bottomrule
\end{tabular}

\end{table}

\subsection*{Models and prompting protocol}

We evaluated fourteen models across six providers: OpenAI, DeepSeek, Qwen, Meta (Llama), Mistral, and Google (MedGemma). We select two models for each provider: a smaller and a larger model variant. For OpenAI and Qwen, whose larger models are of undisclosed size, we also ran GPT-OSS 120B and Qwen 3.5 35B, which are open-weight models that have reported their sizes. Twelve of the fourteen were accessed through the OpenRouter API. The two MedGemma models\cite{sellergren2025medgemma}, which are two medically specialised models, were accessed through a third-party inference endpoint (dr7.ai) that serves the released MedGemma weights, because they were not available through OpenRouter at the time of the runs. We used an identical prompt across all models and conditions, which ensures that observed differences reflect model behavior rather than prompt engineering. Decoding temperature was set at 0 for all models. The full list of models, providers, and settings is provided in Table~\ref{tab:models}. We also provide the prompt used in each setting in the supplementary material (Figure~\ref{fig:supp-prompts}).

\begin{table}[H]
\centering
\small
\caption{Language models evaluated in this study, grouped by provider. Each provider contributes a smaller and a larger model. We report model size, context window size, and ``Evaluated on'' which is the dates on which the runs reported here were executed.}
\label{tab:models}
\resizebox{\linewidth}{!}{%
\begin{tabular}{llcccc}
\toprule
\textbf{Provider} & \textbf{Model} & \textbf{Size (total\,/\,active)} & \textbf{Context} & \textbf{Released} & \textbf{Last Evaluated on (2026)} \\
\midrule
OpenAI & GPT-OSS 20B & 21B\,/\,3.6B & 131K & Aug-2025 & 23 Sep \\
 & GPT-5.6 Luna & undisclosed & 1.05M & Jul-2026 & 23 Sep \\
\addlinespace[2pt]
DeepSeek & DeepSeek V4 Flash & 284B\,/\,13B & 1M & Apr-2026 &  23 Sep \\
 & DeepSeek V4 Pro & 1.6T\,/\,49B & 1M & Apr-2026 &  23 Sep \\
\addlinespace[2pt]
Qwen & Qwen 3.5 9B & 9B & 262K & Mar-2026 & 23 Sep \\
 & Qwen 3.7 Plus & undisclosed & 1M & Jun-2026 & 23 Sep \\
\addlinespace[2pt]
Meta & Llama 3.1 8B & 8B & 128K & Jul-2024 & 23 Sep \\
 & Llama 4 Maverick & 400B\,/\,17B & 1M & Apr-2025 & 23 Sep \\
\addlinespace[2pt]
Mistral & Ministral 8B & 8B & 256K & Dec-2025 & 23 Sep \\
 & Mistral Large 3 & 675B\,/\,41B & 256K & Dec-2025 &  24 Sep \\
\addlinespace[2pt]
Google & MedGemma 4B & 4B & 128K & May-2025 & 23--24 Sep \\
 & MedGemma 27B & 27B & 128K & May-2025 & 23--24 Sep \\
\midrule
\multicolumn{6}{l}{\textit{Additional open-weight models}}\\
OpenAI & GPT-OSS 120B & 117B\,/\,5.1B & 131K & Aug-2025 & 24 Sep \\
Qwen & Qwen 3.5 35B & 35B\,/\,3B & 262K & Feb-2026 & 24 Sep \\
\bottomrule
\end{tabular}}
\end{table}

Because provider-hosted endpoints are updated without notice, we record the exact API model identifier, the decoding parameters, and the request timestamp for every call, and we report results against the checkpoints served on those dates rather than against the model families in general. The per-run manifests deposited with the analysis artifacts contain these identifiers, timestamps, token counts, and costs for every request.

\subsection*{Setting 1: Multiple-choice baseline}
\phantomsection\label{sec:setting1}

In the first setting, each question appears in its original multiple-choice examination format. This is the standard format in question-answering \cite{robinson2023leveraging} and the field has used to track medical competence, and we preserve it unchanged so our results stay comparable with earlier work \cite{lievin2024can}. We use the multiple-choice performance as our main results. Each question appeared with its answer choices, and we scored the extracted choice against the answer key. Per-benchmark accuracy is the fraction of correct questions. The ``Overall'' figure pools all 3,076 questions, weighted by benchmark size. We provide the full model output from this setting in the supplementary material (Figure~\ref{fig:supp-reasoning}).

\subsection*{Setting 2: Generative question answering}
\phantomsection\label{sec:setting2}

We remove the answer choice and require the model to produce the answer itself. In this setting, the model received the question alone, with no choices, and produced a free-text answer. A judge model then compares and decides whether that answer from the model gives the same answer as the ground truth. The judge is a separate model that is not one of the twelve under evaluation, and it is described in the subsequent section. This turns the task from recognition the answer choices into clinical knowledge recall. We evaluate this setting on MedQA-USMLE, MedMCQA and MedBullets, whose ground truths are in medical sentence or word format. We omit this setting for PubMedQA as its gold answers are always \emph{yes}, \emph{no}, or \emph{maybe}.

\subsection*{Setting 3: Question reconstruction as a memorization proxy}
\phantomsection\label{sec:setting3}

We invert the task and show only the answer choices, asking the model to reconstruct the question they belong to. A model that can reconstruct a specific case from its choices alone has likely encountered it before, which makes this a bounded probe of memory rather than of reasoning. To test whether current performance reflects prior exposure to the questions, we added a reconstruction condition. The model saw only a question's answer choices, not the stem, and tried to regenerate a question testing the same knowledge. The judge decided whether the regenerated question matched the original. Reproducing a specific clinical scenario from its choices alone suggests the model has seen the question before. A model seeing it for the first time has little to work from. While we acknowledge that this is not a direct measure of memorization---without knowing the training data there is no direct measure---it serves as a strong bounded proxy to memorization. A model may reconstruct a question by inference from distinctive choices rather than recall, and a failed reconstruction does not prove the question was unseen. We therefore treat reconstruction fidelity as a proxy and hedge conclusions accordingly. Reconstruction applies only to MedQA-USMLE, MedMCQA and MedBullets. PubMedQA's answer contain only \emph{yes}, \emph{no}, or \emph{maybe} choices that are identical across questions, which leaves nothing to distinguish one question from another.

\subsection*{LLM-as-a-judge for generative outputs}
\label{judge}

Free-text medical answers take many correct forms, and as a consequence, methods like string matching cannot score the generative and reconstruction conditions. We implement an LLM-as-a-judge with Claude Sonnet 5 at temperature 0 and provide a single frozen rubric. This judge model is not one of the fourteen models under evaluation. We applied the same rubric and grader to all fourteen models and both conditions, which makes the judge a fixed instrument rather than a tunable part of the analysis. This follows standard LLM-as-judge practice and freezes both rubric and grader to prevent drift. We provide both grading prompts in the supplementary material (Figure~\ref{fig:supp-judge}).


\subsection*{Reconstructable versus un-reconstructable question pools}

Several analyses use the reconstruction verdict at the question level. For each model, we joined its multiple-choice correctness with its reconstruction outcome on the same MedQA-USMLE, MedMCQA and MedBullets questions, then split the questions into that model's reconstructable and un-reconstructable pools. If a model's score rested on memorized questions, its accuracy should fall on the un-reconstructable pool, the questions it cannot reproduce. If the score rests on general competence, accuracy should hold there. Higher accuracy on reconstructable questions, in the other direction, measures the familiarity advantage. Each model is its own control, since the pools are defined per model. Two companion analyses use the same join: reconstruction fidelity split by whether the model answered the question correctly, and, for each provider pair, reconstruction on the questions only the larger model gets right versus that model's baseline rate.

\subsection*{Model size, ensemble, and question-difficulty analyses}

Throughout, accuracy is reported in percentage points and every difference between accuracies is reported in percentage points. After the first mention in each section we shorten this to ``points''. Three further analyses come from the same predictions. For model size, we paired the smaller and larger model in each family and compared them across the three conditions. For ensembling, we combined predictions on each question by majority vote, breaking ties in favour of the highest-scoring member. We did this for every possible committee at every size, within three pools: all six model pairs, the six larger models, and the six smaller models. At each size we report the best-scoring committee, which is selected with hindsight on the same questions and therefore bounds what any committee-selection procedure could achieve. Each committee is compared against the best single model in its own pool. For question difficulty, we counted how many of the twelve models answered each question correctly across all four benchmarks and examined the two extremes: questions all twelve solved and questions all twelve missed. For the missed questions we recorded the consensus answer and flagged those where at least eight of the twelve agreed on a single non-key answer, then inspected the flagged questions for recurring patterns. We report the full analysis results in the subsequent section.

\section*{RESULTS}

In this section we analyze the results following the four research questions and close with three additional analyses. We include the full numerical results, prompt templates, reasoning traces, and judge results in the supplementary material.

\subsection*{Finding 1: Current reasoning models reach expert-level performance on two medical benchmarks}

We evaluated fourteen LLMs on all 3,076 medical questions: the twelve models of the main roster and two open-weight models that explicitly reported their sizes (GPT-OSS 120B and Qwen 3.5 35B). Because GPT-5.6 Luna and Qwen 3.7 Plus are closed models of undisclosed size, we added these open-weight models from the same providers, which are larger than GPT-OSS 20B and Qwen 3.5 9B in total parameters, to validate the comparison between smaller and larger models of open-weight models. Every model surpassed the MedQA-USMLE human passing score of 60\% (Table~\ref{tab:rq1}). Accuracy ran from 67.5\% to 94.7\% on MedQA-USMLE, from 58.4\% to 83.8\% on MedMCQA, and from 74.8\% to 81.6\% on PubMedQA. MedBullets was harder for every model: accuracy there ran from 57.4\% to 89.1\%, and each model scored 5.5 to 16.2 points below its own MedQA-USMLE result, by 7.7 points on average for the six larger models and 12.3 for the six smaller ones. Qwen 3.7 Plus recorded the highest score on MedQA-USMLE, MedMCQA and PubMedQA (94.7\%, 83.8\%, 81.6\%), with GPT-5.6 Luna matching it on the first two and scoring highest on MedBullets (89.1\%, against 87.8\% for Qwen 3.7 Plus). The two medically specialised models (MedGemma 4B and 27B) perform competitively with general-purpose models: MedGemma 27B achieves 84.7\% overall and MedGemma 4B achieves 78.1\%. MedGemma 4B, despite being the smallest model, outperforms most smaller models from other provider (three of five), while MedGemma 27B, with fewer parameters than most larger models, still surpasses two larger models (Llama 4 and Mistral Large 3). The two added open-weight models fall in the middle of the roster, at 84.3\% (Qwen 3.5 35B) and 83.1\% (GPT-OSS 120B) overall. Each scores above the smaller model of its provider and below the provider's closed flagship model. The confidence intervals qualify that ranking: the four leading models---Qwen 3.7 Plus, GPT-5.6 Luna, DeepSeek V4 Pro, and DeepSeek V4 Flash---have overlapping intervals on every benchmark, which leaves them unseparated at these sample sizes. The leaderboard resolves model tiers, not individual positions within the leading tier.

\begin{table}[H]
\centering
\small
\caption{Multiple-choice accuracy (\%) on four benchmarks, reported as accuracy followed by its 95\% confidence interval in brackets. Current models (2026) are ordered by their mean accuracy across all 3,076 questions. The lower block reports prior-generation models and human baselines from prior work \cite{lievin2024can}, and no intervals are computed for them. The best current-model value in each column is in bold. \protect\mbox{$^{\dagger}$}Open-weight models added to compare smaller and larger models whose sizes are known.}
\label{tab:rq1}
\resizebox{\linewidth}{!}{%
\begin{tabular}{lcccc}
\toprule
\textbf{Model} & \textbf{MedQA-USMLE} & \textbf{MedMCQA} & \textbf{PubMedQA} & \textbf{MedBullets}\\
\midrule
\multicolumn{5}{l}{\textit{This work: current models (2026)}}\\
\addlinespace[1pt]
Qwen 3.7 Plus & \textbf{94.66} [93.3, 95.8] & \textbf{83.80} [81.4, 86.0] & \textbf{81.60} [78.0, 84.8] & 87.79 [83.6, 91.0] \\
GPT-5.6 Luna & \textbf{94.66} [93.3, 95.8] & \textbf{83.80} [81.4, 86.0] & 78.00 [74.2, 81.4] & \textbf{89.11} [85.1, 92.1] \\
DeepSeek V4 Pro & 94.42 [93.0, 95.6] & 82.40 [79.9, 84.6] & 76.00 [72.1, 79.5] & 87.46 [83.3, 90.7] \\
DeepSeek V4 Flash & 93.95 [92.5, 95.1] & 81.80 [79.3, 84.1] & 77.80 [74.0, 81.2] & 88.45 [84.4, 91.6] \\
MedGemma 27B & 91.99 [90.4, 93.4] & 79.70 [77.1, 82.1] & 77.40 [73.5, 80.8] & 82.51 [77.8, 86.4] \\
Qwen 3.5 35B\mbox{$^{\dagger}$} & 91.91 [90.3, 93.3] & 77.30 [74.6, 79.8] & 80.60 [76.9, 83.8] & 81.19 [76.4, 85.2] \\
Llama 4 Maverick & 89.87 [88.1, 91.4] & 79.40 [76.8, 81.8] & 80.80 [77.1, 84.0] & 79.21 [74.3, 83.4] \\
Mistral Large 3 & 88.77 [86.9, 90.4] & 80.80 [78.2, 83.1] & 77.40 [73.5, 80.8] & 82.18 [77.5, 86.1] \\
GPT-OSS 120B\mbox{$^{\dagger}$} & 90.57 [88.8, 92.1] & 75.50 [72.7, 78.1] & 78.40 [74.6, 81.8] & 84.16 [79.6, 87.8] \\
Qwen 3.5 9B & 91.91 [90.3, 93.3] & 74.10 [71.3, 76.7] & 80.80 [77.1, 84.0] & 77.23 [72.2, 81.6] \\
MedGemma 4B & 85.55 [83.5, 87.4] & 73.00 [70.2, 75.7] & 74.80 [70.8, 78.4] & 69.31 [63.9, 74.2] \\
GPT-OSS 20B & 85.31 [83.3, 87.1] & 68.80 [65.9, 71.6] & 75.00 [71.0, 78.6] & 72.28 [67.0, 77.0] \\
Ministral 8B & 77.38 [75.0, 79.6] & 69.30 [66.4, 72.1] & 77.00 [73.1, 80.5] & 63.04 [57.5, 68.3] \\
Llama 3.1 8B & 67.48 [64.9, 70.0] & 58.40 [55.3, 61.4] & 77.80 [74.0, 81.2] & 57.43 [51.8, 62.9] \\
\midrule
\multicolumn{5}{l}{\textit{Prior results and human baselines}}\\
\addlinespace[1pt]
Human (expert) & 87.00 & 90.00 & 78.00 & -- \\
GPT-4 (2023) & 86.10 & 73.70 & 81.20 & -- \\
Med-PaLM 2 (2023) & 86.50 & 72.30 & 77.40 & -- \\
Llama 2 70B CoT (2023) & 62.50 & 53.60 & -- & -- \\
Codex 5-shot CoT (2022) & 60.20 & 59.70 & 78.20 & -- \\
Human (passing) & 60.00 & 50.00 & -- & -- \\
Fine-tuned SOTA (2022) & 50.30 & 52.90 & 78.20 & -- \\
\bottomrule
\end{tabular}}
\end{table}

Measured against the human-expert baselines, the strongest model clears the expert bar on MedQA-USMLE and PubMedQA but not on MedMCQA. It exceeds the 87.0\% expert score on MedQA-USMLE by 7.7 percentage points, comfortably outside the interval, and the 78.0\% expert score on PubMedQA by 3.6 points, though there the expert value sits at the lower bound of the interval [78.0, 84.8] and the advantage is marginal. On MedMCQA it falls 6.2 points short of the 90.0\% expert score, a gap wider than the interval. Expert-level performance is therefore a firm claim only on the licensing-examination benchmark, a marginal one on PubMedQA, and false on MedMCQA. MedBullets has no published human baseline, so it does not enter this comparison.

Measured against the previous model generation, the gain is concentrated on MedQA-USMLE. Ten of the fourteen models exceeded the 2023 GPT-4 reference of 86.1\%, and even a 9B model, Qwen 3.5 9B, beat both GPT-4 and Med-PaLM 2 there. Against Codex 5-shot chain-of-thought in 2022, the best MedQA-USMLE score rose from 60.2\% to 94.7\%, a gain of 34.5 points. MedMCQA and PubMedQA have not moved the same way. On MedMCQA the best score improves on the strongest 2023 model by 10.1 points yet still trails expert performance. PubMedQA is flatter still: the best 2026 score of 81.6\% barely exceeds GPT-4's 81.2\% from 2023, leaving almost no headroom beyond the earlier frontier. Saturation is therefore a property of the licensing-exam benchmark alone.

From Finding 2 onward, we report the twelve models of the main roster. The two added models follow the same trends with a smaller margin when they are used in place of the two flagship models, GPT-5.6 Luna and Qwen 3.7 Plus. The full results are in Supplementary Material S16 and Tables S1--S6.

\subsection*{Finding 2: Smaller models trail their larger counterparts by 8.7 percentage points}

\begin{figure}[H]
\centering
\includegraphics[width=0.95\linewidth]{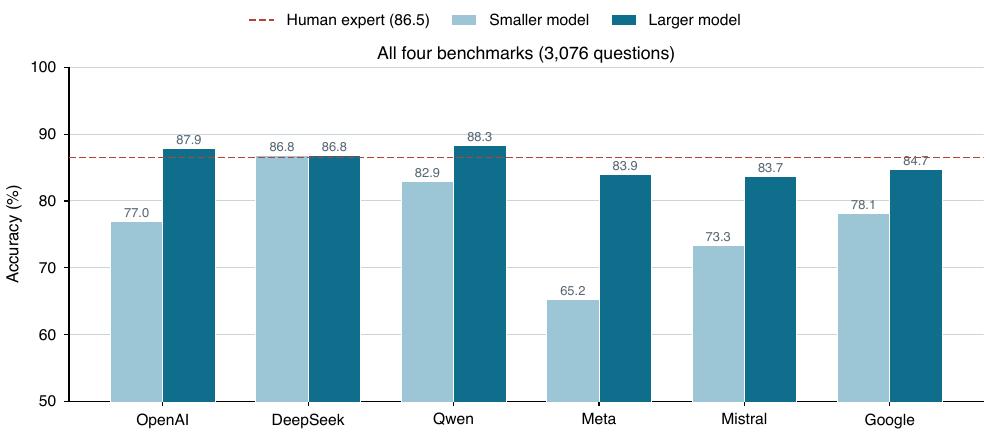}
\caption{Multiple-choice accuracy on all 3,076 questions of the four benchmarks, with each provider's smaller and larger model side by side. The dashed line is the human-expert reference of 86.5\%, weighted by benchmark size over MedQA-USMLE, MedMCQA and PubMedQA, the benchmarks that have one; Table~\ref{tab:rq1} gives each benchmark separately.}
\label{fig:main1}
\end{figure}

We look at the performance for each pair of LLMs from the same provider. Pairing each provider's smaller and larger model isolates the effect of size while holding training recipe and data roughly constant (Figure~\ref{fig:main1}). The larger model wins in nearly every pair, by 8.7 percentage points on average across all 3,076 questions (six provider pairs, paired \mbox{$t$}-test \mbox{$t(5)=3.37$}, \mbox{$p=.020$}). The advantage is uneven across benchmarks: it reaches 8.8 points on MedQA-USMLE and 10.8 on MedMCQA, but narrows to 1.3 points on PubMedQA, where DeepSeek's smaller model in fact outscores its larger sibling (77.8\% versus 76.0\%). On MedBullets the advantage is the largest of the four benchmarks, 13.4 points (\mbox{$t(5)=4.05$}, \mbox{$p=.010$}), and the DeepSeek pair is again the exception, its smaller model 1.0 point ahead.

Size also decides who clears the expert bar. On MedQA-USMLE all six larger models exceed the human-expert baseline of 87.0\%, whereas only two of the six smaller models do. The generational gain is therefore broad rather than the work of a single laboratory, but within each provider it is concentrated in the larger model.

The spread among the smaller models is the more consequential result. On MedQA-USMLE the smaller models range from 67.5\% to 94.0\% while the larger ones fall within 5.9 points of one another, and MedMCQA shows the same pattern (a 23.4-point spread among smaller models against 4.4 among larger), and MedBullets shows it most strongly (31.0 against 9.9). Qwen 3.5 9B reaches 91.9\%, within 2.8 points of its own frontier sibling, whereas Llama 3.1 8B trails at 67.5\%. The 24.4-point gap between those two smaller models is wider than the gap between any smaller model and its own larger sibling. For a team restricted to models it can host itself, which smaller model it selects therefore matters more than whether a frontier model is out of reach. One caveat qualifies the pairing: DeepSeek V4 Flash is a speed-optimized variant rather than a substantially smaller model, which is why that pair shows the narrowest gap throughout. We provide this comparison across all three settings in the supplementary material (Table~\ref{tab:rq2scale}), together with an example of one provider's smaller and larger model answering the same question (Figure~\ref{fig:supp-paired}). With the two added open-weight models in place of GPT-5.6 Luna and Qwen 3.7 Plus (Finding 1), the difference between smaller and larger models still holds, with a smaller margin: across the six provider pairs, the larger model leads by 7.2 points on average instead of 8.7 (Supplementary Material S16).

\subsection*{Finding 3: Removing answer choices cuts accuracy by 31.3 percentage points}

\begin{figure}[H]
\centering
\includegraphics[width=0.95\linewidth]{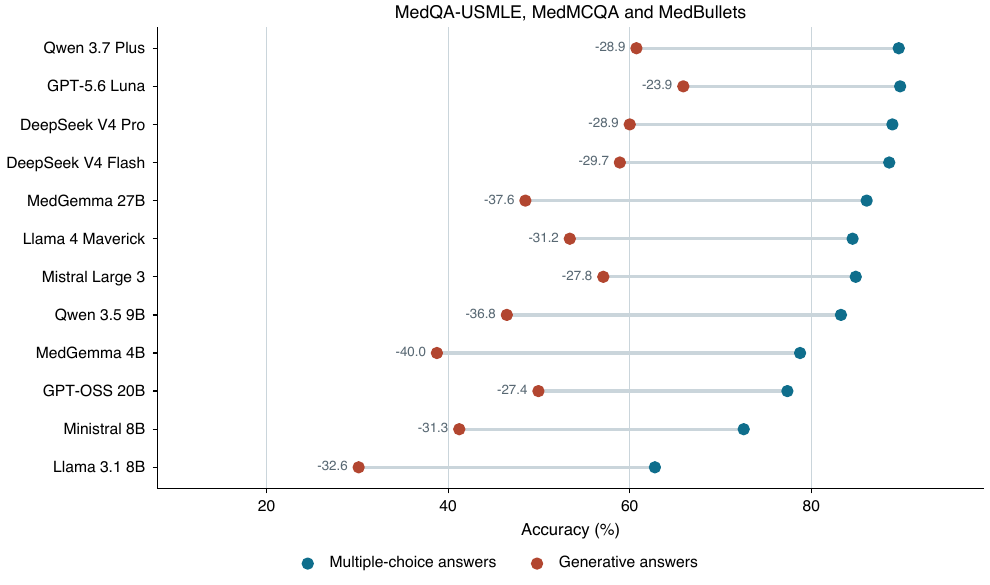}
\caption{Accuracy on MedQA-USMLE, MedMCQA and MedBullets pooled (2,576 questions), comparing multiple choice (setting 1) and generative setting (setting 2), one row per model. Teal markers are multiple-choice accuracy, red markers the generative setting. The connecting segment is the difference between them, given in points.}
\label{fig:main2}
\end{figure}

In the generative question-answering condition (\hyperref[sec:setting2]{Setting 2}), we ask LLMs the same questions as in \hyperref[sec:setting1]{Setting 1}, but we withhold the choices and grade the open-ended answer, on MedQA-USMLE, MedMCQA and MedBullets. The mean fell from 82.3\% to 50.9\%, a drop of 31.3 percentage points (Figure~\ref{fig:main2}). To check that this drop is not an accident of which models we happened to evaluate, we compared each model against itself across the two conditions with a paired $t$-test over the twelve models. The drop is far larger than the variation between models (\mbox{$t(11)=23.1$}, $p<.0001$), meaning a difference this size would essentially never arise by chance. Testing each model on its own questions with McNemar's test, which compares paired correct/incorrect outcomes on the same questions, the drop is significant for every one of the twelve individually ($p<.001$ in each case).

The effect was large but not uniform, ranging from roughly 24 to 40 points. The two largest drops in the cohort belong to the two medically specialised models: MedGemma 4B loses 40.0 points and MedGemma 27B 37.6, against a previous maximum of 36.8. This suggests that models that are trained or fine-tuned on medical data rely more on the answer choices than general models. On MedBullets alone the ordering differs. Qwen 3.5 9B (35.3 points) and Llama 3.1 8B (34.7) drop the most, then MedGemma 27B (32.3), while MedGemma 4B drops 28.4, close to the average (Table~\ref{tab:rq4}). GPT-5.6 Luna stayed the strongest open-ended model, but its score still fell sharply. This is a key finding of our experiments. It implies that in multiple-choice settings, LLMs can pass medical benchmarks, but when answer choices are not available, the performance drops significantly.

PubMedQA is not included in this finding as its answers are always \emph{yes}, \emph{no}, or \emph{maybe}, and as a consequence, hiding the choices does not make the task open-ended. The model still makes the same three-way judgment. Including it shrinks the pooled drop from 31.3 to 25.4 points, understating the effect. Table~\ref{tab:rq4} therefore reports MedQA-USMLE, MedMCQA and MedBullets, the benchmarks the claim rests on, and its ``Overall'' columns match the 31.3-point headline.

The gap grows when the answer choices are withheld. The larger model of each provider leads by 10.1 points with the choices and by 13.4 points without them, and the gap widens in all six provider pairs (paired \mbox{$t$}-test, \mbox{$t(5)=3.16$}, \mbox{$p=.025$}). Multiple-choice setting therefore make the smaller models look closer to their larger counterparts than when they are tested without choices.

\subsection*{Finding 4: Models can reconstruct 25\% of the questions from the answer choices alone}

\begin{figure}[!htb]
\centering
\includegraphics[width=0.95\linewidth]{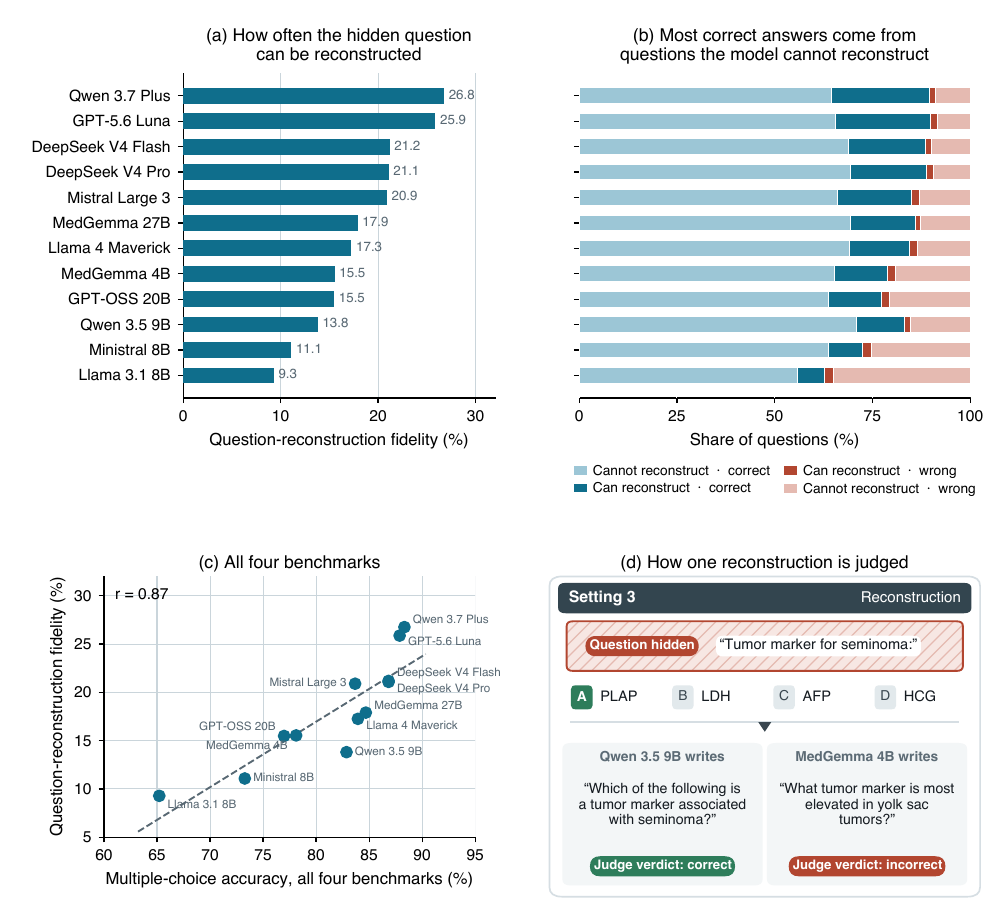}
\caption{Question reconstruction (Setting 3). (a) How often each model reconstructs the hidden question from the answer choices alone. (b) Every MedQA-USMLE, MedMCQA and MedBullets question resolved four ways, by whether the model reconstructed it and whether it answered it correctly. (c) Reconstruction rate against multiple-choice accuracy on all four benchmarks, one point per model; the dashed line is the least-squares fit. (d) How one reconstruction is judged, on a real MedMCQA question: the model sees only the answer choices, and the judge compares its question with the hidden original.}
\label{fig:main3}
\end{figure}

LLMs may perform well simply because they have seen the questions during training. We probe this with question reconstruction (\hyperref[sec:setting3]{Setting 3}), and the rates are not trivial: top models reconstruct roughly a quarter of the hidden questions from the answer choices alone---26.8\% for Qwen 3.7 Plus and 25.9\% for GPT-5.6 Luna, falling to 9.3\% for the weakest model, with the two MedGemma models at the middle at 17.9\% and 15.5\% (Figure~\ref{fig:main3}a; Table~\ref{tab:rq3}). MedBullets gives similar rates: 29.4\% for Qwen 3.7 Plus and 28.1\% for GPT-5.6 Luna, falling to 10.6\% for Llama 3.1 8B. A specific clinical question cannot be guessed by chance, which makes a quarter a substantial hit rate. It raises the obvious worry that these models hold question and answer together in memory and are retrieving rather than reasoning.

Reconstruction tracks capability closely. Plotting each model's reconstruction rate against its multiple-choice accuracy gives a strong positive relationship (Figure~\ref{fig:main3}c): the Pearson correlation across all four benchmarks is 0.87 on a scale where 1.0 would be a perfect straight line and a relationship this strong would arise by chance fewer than once in a thousand samples (\mbox{$p<.001$}). The models that answer best are also the models that have most evidently encountered these questions before, and the ordering is nearly the same in both. That is exactly the pattern a contamination account predicts, which is why we test it directly rather than dismiss it. It is also, however, what one expects if larger and better-trained models are simply better at everything, including inferring a plausible question from a distinctive set of choices. The correlation establishes that the two travel together. It cannot say which drives which.
The rate alone therefore cannot settle the worry in either direction, because a model may reconstruct a question from memory or by inferring it from distinctive choices, and both produce a match. What reconstruction measures is familiarity with a question, not the use of memory to answer it. Familiarity does travel with success at the question level. Every model reconstructs more often on the questions it answered correctly than on those it got wrong, by 7.8 points on average (Table~\ref{tab:rq9}). Easier questions are plausibly easier both to answer and to reconstruct, and as a consequence, this shows the proxy carries signal rather than that memory drives the score. The composition of each model's answers is more decisive (Figure~\ref{fig:main3}b). Resolving every MedQA-USMLE, MedMCQA and MedBullets question by whether the model reconstructed it and whether it answered correctly, the largest category for every model is the same: questions answered correctly that it could not reconstruct, running from 55.9\% of all questions for Llama 3.1 8B to 71.0\% for Qwen 3.5 9B. The questions a model both reconstructed and answered correctly---the only ones a memorization account can claim---account for between 6.9\% and 25.2\%. Whatever familiarity contributes, it is not carrying the bulk of the score.

We ran additional verification on Setting 3. First, we tested the judge on questions where we knew the correct verdict in advance. We perturb the question by constructing a positive pair (an original benchmark question and a paraphrase of it) and a negative pair (a reconstructed question and randomly chosen answer choices). The judge accepted 119 of the 120 positive pairs and rejected all 156 negative pairs. Second, we evaluated the judge against human and NLP measures, described in \mbox{\hyperref[sec:validation-judge]{Validation of the LLM judge}}. Its verdicts on reconstructions follow their text similarity to the original question, and an independent expert agreed with 49 of its 50 verdicts on generative answers and questions. Together, these verifications show that when providing the ground truth with two candidate answers, the judge is accurate; hence, the reconstruction rates are not inflated. Further validations, including an inspection of accepted reconstructions and a negative control with newly written questions are presented in Supplementary Material S13.

\subsection*{Finding 5: LLMs perform significantly better on questions they can reconstruct}

\begin{figure}[H]
\centering
\includegraphics[width=0.95\linewidth]{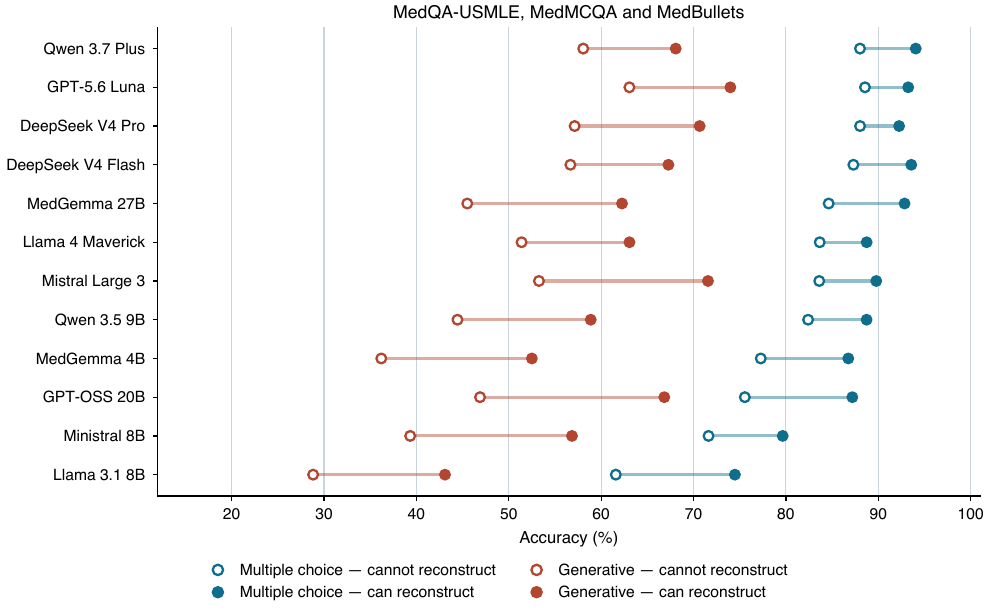}
\caption{Accuracy on the questions each model can and cannot reconstruct, in the multiple-choice and generative settings. Filled markers are questions the model reconstructed, hollow markers those it could not. The length of each segment is the difference between the two pools. Questions from MedQA-USMLE, MedMCQA and MedBullets are pooled.}
\label{fig:main4}
\end{figure}

One of the hypotheses we have is if benchmark performance were carried by memorized questions, accuracy should fall on the questions a model cannot reproduce. As shown in Figure~\ref{fig:main4} and  Table~\ref{tab:rq5} the memorization effect does persist, and its size depends on the format. In multiple choice, every model scores higher on reconstructable  than un-reconstructable questions, by 7.4 percentage points on average. Once the choices are withheld this effect roughly doubles, to 14.5 points. Comparing each model against itself with a paired $t$-test over the twelve models, both gaps are far larger than the variation between models and are highly unlikely to have arisen by chance (\mbox{$t(11)=9.41$} and \mbox{$t(11)=15.33$} respectively, $p<.0001$ in both cases). Figure~\ref{fig:main4} shows the difference directly: the gap in performance from each model in generative segment (red) is visibly larger than its multiple-choice segment (blue).

A third comparison points the same way, though less firmly. For each provider pair we isolated the questions that only the larger model answers correctly and asked whether they are more reconstructable than that model's questions in general. They are not: reconstruction on those unique wins falls below the larger model's own baseline in all six providers, by a mean of 6.5 points, a difference unlikely to be chance across the six pairs (paired $t$-test, \mbox{$t(5)=-6.47$, \mbox{$p=.001$}}). The larger model's advantage therefore sits on questions it is \emph{less} familiar with. The DeepSeek estimate rests on only 90 questions and is individually within sampling noise, and we therefore treat this as supporting rather than primary evidence.


\subsection*{Finding 6: Smaller models give the most accuracy per dollar, but they remain below human expert performance}

\begin{figure}[!htb]
\centering
\includegraphics[width=0.95\linewidth]{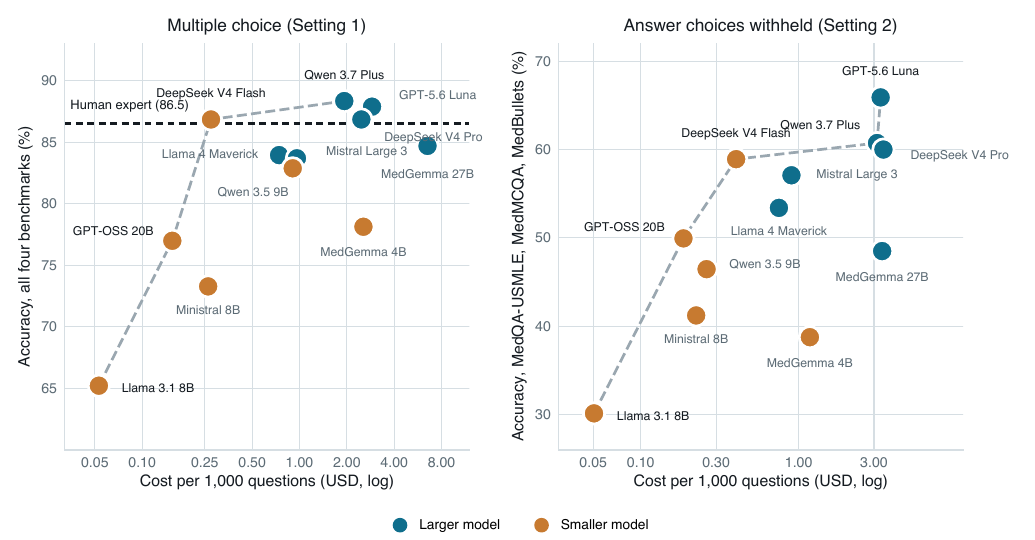}
\caption{Accuracy against inference cost per thousand questions, one point per model, on a logarithmic cost axis. (A) Multiple choice, on all four benchmarks. (B) Answer choices withheld, on MedQA-USMLE, MedMCQA and MedBullets, with Setting 2 cost on an equal-prompt basis. The dashed line joins the models on the cost-accuracy frontier. The horizontal line in (A) is the human-expert baseline weighted by benchmark size over MedQA-USMLE, MedMCQA and PubMedQA, the benchmarks that have one.}
\label{fig:cost-benefit}
\end{figure}

Benchmark accuracy is usually reported without its price. Because every request here was recorded, we can pair the two directly (Table~\ref{tab:cost}; Figure~\ref{fig:cost-benefit}). Accuracy and cost order the models differently. First, DeepSeek V4 Flash answers at 86.8\%, 1.5 percentage points below the highest-scoring model, for one-seventh of its price. The cost comparison covers the twelve models billed through two API providers. Per thousand correct answers, GPT-5.6 Luna costs \$3.29 against DeepSeek V4 Flash's \$0.32---a ten times premium for one point of accuracy. Second, the cost-accuracy frontier is short and steeply diminishing. Only four of the twelve models lie on it, three of them smaller models, and accuracy climbs 22 points between \$0.05 and \$0.27 per thousand questions but only 1.5 further points over the next seven-fold rise in price. Third, cheapness is not a substitute for competence. Weighting the expert baselines of MedQA-USMLE, MedMCQA and PubMedQA by benchmark size gives a combined reference of 86.5\%, which GPT-OSS 20B misses by 9.5 points and Llama 3.1 8B by 21.3. Among the smaller models only DeepSeek V4 Flash reaches it. The conclusion is therefore narrower than ``small is enough'': accuracy per dollar favours the smaller tier, but most of that tier remains well short of expert performance.

Note that these are provider list prices for hosted inference on the dates of our runs. They are price from calling a model rather than self-hosting one, which is the mode that makes smaller open-weight models more accessible in the first place, and prices move faster than capabilities. 

\begin{table}[!htb]
\centering
\small
\caption{Recorded inference cost in both answering settings, ordered by multiple-choice accuracy. Setting 1 covers all 3,076 questions of the four benchmarks, and Setting 2 covers the 2,576 MedQA-USMLE, MedMCQA and MedBullets questions. Tokens are prompt plus completion. The final column of each block divides cost by accuracy, giving the price of obtaining a thousand correct answers. \mbox{$^{\dagger}$} The two MedGemma models were served through a different provider (Dr7.ai).}
\label{tab:cost}
\resizebox{\linewidth}{!}{%
\begin{tabular}{lcccccc}
\toprule
\multirow{2}{*}{\textbf{Model}} & \multicolumn{3}{c}{\textbf{Setting 1: multiple choice}} & \multicolumn{3}{c}{\textbf{Setting 2: choices withheld}} \\
\cmidrule(lr){2-4}\cmidrule(lr){5-7}
 & Tokens & Cost (\$) & Cost per 1,000 & Tokens & Cost (\$) & Cost per 1,000 \\
 & & & correct answers (\$) & & & correct answers (\$) \\
\midrule
Qwen 3.7 Plus & 8.06M & 5.92 & 2.18 & 5.03M & 5.26 & 3.36 \\
GPT-5.6 Luna & 4.97M & 8.89 & 3.29 & 1.26M & 3.80 & 2.24 \\
DeepSeek V4 Pro & 5.80M & 7.60 & 2.85 & 2.55M & 4.84 & 3.13 \\
DeepSeek V4 Flash & 6.00M & 0.84 & \textbf{0.32} & 3.26M & 0.65 & 0.43 \\
MedGemma 27B$^{\dagger}$ & 4.97M & 20.03 & 7.69 & 0.46M & 1.54 & 1.23 \\
Llama 4 Maverick & 5.59M & 2.28 & 0.88 & 1.47M & 0.80 & 0.58 \\
Mistral Large 3 & 5.17M & 2.96 & 1.15 & 1.03M & 0.71 & 0.48 \\
Qwen 3.5 9B & 17.63M & 2.78 & 1.09 & 1.37M & 0.17 & \textbf{0.14} \\
MedGemma 4B$^{\dagger}$ & 5.56M & 7.84 & 3.26 & 0.53M & 0.65 & 0.65 \\
GPT-OSS 20B & 5.82M & 0.48 & \textbf{0.20} & 2.33M & 0.27 & 0.21 \\
Ministral 8B & 5.75M & 0.81 & 0.36 & 1.09M & 0.14 & \textbf{0.13} \\
Llama 3.1 8B & 5.19M & 0.16 & \textbf{0.08} & 1.01M & 0.02 & \textbf{0.03} \\
\midrule
\textit{Total} & 80.5M & \textit{60.60} & & 21.4M & \textit{18.85} & \\
\bottomrule
\end{tabular}}
\end{table}

\subsection*{Finding 7: LLMs ensemble does not beat the best single model, and multiplies the cost}

A majority vote across all twelve models reached 88.0\%, which is indistinguishable from the best single model at 88.3\% (McNemar's test, \mbox{$p=.47$}; Table~\ref{tab:ensemble}). This much is expected: when a strong model votes with weaker ones, the weaker majority can outvote it wherever they disagree. The more useful question is how much any committee could gain, and at what cost.

\begin{figure}[H]
\centering
\includegraphics[width=0.98\linewidth]{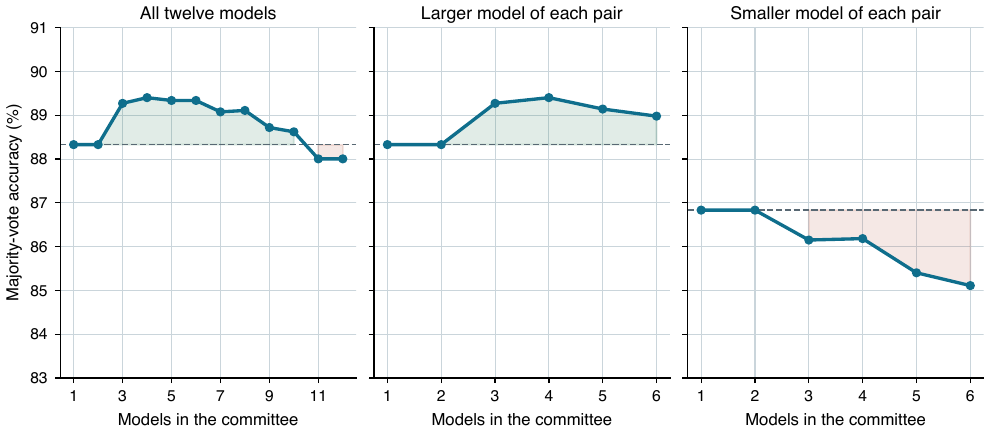}
\caption{Majority-vote accuracy against committee size. At each size \mbox{$k$} the best-scoring of the possible committees is shown. The dashed line in each panel is that pool's own best single model; shading marks where voting gains on it and where it loses. All 3,076 questions of the four benchmarks are pooled.}
\label{fig:ensemble}
\end{figure}

Even a committee chosen with hindsight gains little (Figure~\ref{fig:ensemble}). Taking the best of the \mbox{$\binom{12}{k}$} possible committees at each size \mbox{$k$}, which no deployable procedure could beat, the vote peaks at 89.4\% with four models, 1.1 points above the best single model, and a committee of larger models alone reaches the same 89.4\%. Voting among the six smaller models never beats the best of them: all six together reach 85.1\% against 86.8\% for DeepSeek V4 Flash. The gains that do exist are expensive, because every member answers every question: polling all twelve costs 10.2 times as much as the best single model for 0.3 points less accuracy, and the best four-model committee buys its 1.1 points at 3.1 times the price. Several models are better used for routing and verification, sending each question to the model most likely to answer it, or consulting a second model only when the first is uncertain.

\subsection*{Finding 8: Every model answers 47\% of the questions correctly, but 3.5\% of the questions remain difficult for all models}

\begin{figure}[H]
\centering
\includegraphics[width=0.9\linewidth]{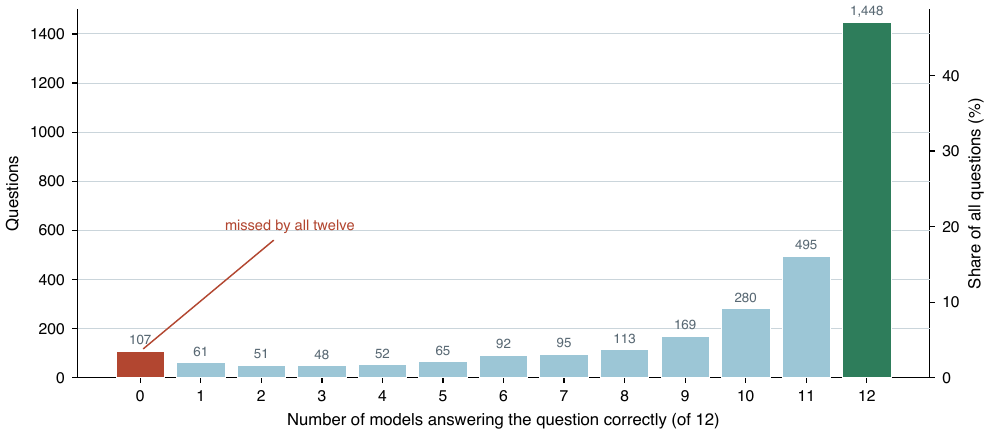}
\caption{Distribution of question difficulty across the 3,076 questions of the four benchmarks. Each bar shows a number of questions that the twelve models answered correctly. The right-hand axis represents the same bars as a percentage of the full question set.}
\label{fig:main5}
\end{figure}

Counting how many of the twelve models answer each question correctly shows a sharply bimodal distribution (Table~\ref{tab:rq8}; Figure~\ref{fig:main5}). All twelve models answered 1,448 of the 3,076 questions in the four benchmarks correctly (47.1\%), while only 107 (3.5\%) were missed by every model. Nearly half of the questions are now solved by every model we tested, including the weakest, and the questions that separate models occupy a narrow band between those two extremes. MedBullets stands apart at the top end: all twelve solve only 99 of its 303 questions (33\%), the lowest share of the four benchmarks, while only 5 (1.7\%) defeat all twelve.

\begin{table}[H]
\centering
\small
\caption{Question difficulty by benchmark: questions solved by all twelve models versus missed by all twelve.}
\label{tab:rq8}
\begin{tabular}{lccc}
\toprule
\textbf{Benchmark} & \textbf{Questions} & \textbf{Solved by all 12} & \textbf{Missed by all 12} \\
\midrule
MedQA-USMLE & 1,273 & 668 (52\%) & 12 (0.9\%) \\
MedMCQA & 1,000 & 395 (40\%) & 48 (4.8\%) \\
PubMedQA & 500 & 286 (57\%) & 42 (8.4\%) \\
MedBullets & 303 & 99 (33\%) & 5 (1.7\%) \\
\midrule
Overall & 3,076 & 1,448 (47\%) & 107 (3.5\%) \\
\bottomrule
\end{tabular}
\end{table}

The small unsolved tail has a distinctive property. Of the 107 questions missed by all twelve, 94 (88\%) drew at least eight models onto a single answer other than the recorded key: the models did not simply fail, they agreed with one another on a different answer. These questions are unevenly distributed, concentrating in MedMCQA (48) and PubMedQA (42) with only 12 in MedQA-USMLE and 5 in MedBullets, an imbalance larger than the differing sizes of the four benchmarks would produce (chi-squared test, \mbox{$\chi^2=66.3$}, $p<.0001$). Inspecting them, three recurring situations account for most (Table~\ref{tab:rq8examples}). The first is MedMCQA questions whose choices contain typographical corruption. The second is questions where the most natural reading points to the answer the models chose rather than the recorded key. The third is PubMedQA questions turning on whether a qualified finding counts as \emph{yes} or \emph{maybe}.

We report this pattern as an observation rather than a verdict on the benchmarks. Convergence among models from six providers is at least consistent with a question admitting a defensible reading other than the key, but we did not obtain independent clinician adjudication, and models trained on overlapping web data can also share a mistake. We therefore describe these questions as warranting review, not as confirmed errors, and release the full set so that others can examine them. We provide three of these questions, one for each pattern described above, in the supplementary material (Figure~\ref{fig:supp-hard}).

What the tail does establish is a practical limit on resolution. Because all twelve models miss these 107 questions, they lower every score equally and cannot change the ranking. They simply place the attainable maximum about 3.5 points below 100\%. The questions that decide close comparisons are the ones some models answer and others miss. Since the leading models in Table~\ref{tab:rq1} are separated by well under one point, differences of that size should be treated with caution until question quality has been examined at comparable precision. Systematic re-examination of labels in general-domain benchmarks has reached similar conclusions\cite{northcutt2021pervasive,gema2024done}.

Additionally, we audited the answer keys of the most difficult questions across all benchmarks with a newer language model (Claude Opus 5.5). It judged the answer key as wrong, outdated, ambiguous for 65 of the 107 questions that every model missed, and for 60 of the 445 questions on which the four leading models disagree. However, removing all 125 ambiguous questions from the analysis pool leaves the performance order of the twelve models unchanged. The full details are provided in Supplementary Material S14

\begin{table}[H]
\centering
\small
\caption{Example questions missed by all twelve models.}
\label{tab:rq8examples}
\begin{tabular}{p{0.13\linewidth}p{0.47\linewidth}cc}
\toprule
\textbf{Benchmark} & \textbf{Question (abbreviated)} & \textbf{Key} & \textbf{Consensus} \\
\midrule
MedMCQA & ``In mandibular primary second molar the true statement is\ldots'', with choices including ``All buccal cusp are fo same size$>$ all lingual cusp are of same size'' & B & A (10/12) \\
\addlinespace[3pt]
PubMedQA & ``Longer-term quality of life and outcome in stroke patients: is the Barthel index \emph{alone} an adequate measure of outcome?'' & yes & no (12/12) \\
\addlinespace[3pt]
PubMedQA & ``Antral follicle assessment as a tool for predicting outcome in IVF---is it a better predictor than age and FSH?'' & maybe & yes (12/12) \\
\addlinespace[3pt]
MedBullets & ``A 43-year-old man presents to the emergency department with bright red blood in his stool this morning\mbox{\ldots} admits having `too much to drink' and vomited profusely overnight\mbox{\ldots} Which of the following is the most likely cause of this patient's symptoms?'' & C & B (12/12) \\
\addlinespace[3pt]
MedBullets & ``A 9-hour-old newborn girl is found in the newborn nursery with diffuse swelling of the scalp not present at birth\mbox{\ldots} This patient's condition most likely affects which of the following spaces or potential spaces?'' & C & B (8/12) \\
\bottomrule
\end{tabular}
\end{table}

\section*{VALIDATION OF THE LLM JUDGE}
\phantomsection\label{sec:validation-judge}

Settings 2 and 3 are scored by a frontier language model, so we validate judge verdicts against two measures using NLP embedding similarity, and against a clinician.

\subsection*{Agreement with text-similarity measures}

We scored all 30,912 outputs in each setting on the 2,576 MedQA-USMLE, MedMCQA and MedBullets questions. Word overlap is the share of the reference's words that appear in the output, with each word weighted by how rare it is in the benchmark. Embedding similarity is defined as the cosine similarity between sentence embeddings of the two texts (all-MiniLM-L6-v2\cite{reimers2019sentence}). Agreement with the judge is summarised by the AUC: the probability that an output that the judge accepted scores higher than one it rejected, with 0.5 indicating no relation.

For generative answers, we compute the correlation between the judge verdicts and the embedding similarity (Table~\ref{tab:judge-similarity}). We analyze two groups of the generative answers. For the answers that the LLM-judge flagged as correct, the generated answers contained 71\% of the answers, on average, from the ground truth. Also, the mean embedding similarity between the generated answers and the ground truth is 0.695. Compared to the second group, where the LLM-judge flagged the answer as incorrect, the generated answers contained 20.8\% of the answers from the ground truth, on average, with the mean embedding similarity dropping to 0.284.


For question reconstruction (Setting 3), we see lower embedding similarity and word overlap, as expected, because the question that the LLM had to reconstruct is relatively longer than the generative answer (Setting 2). Reconstructions that the LLM-judge flagged as correct have a mean embedding similarity of 0.571 and, on average, recovered 18.5\% of the words from the original question, against 0.431 mean embedding similarity and 10.9\% mean word overlap for those that the judge flagged as incorrect.

\begin{table}[H]
\centering
\small
\caption{Agreement between the LLM judge and two text-similarity measures (word overlap percentange and embedding similarity), over the 30,912 outputs in settings 2 and 3.}
\label{tab:judge-similarity}
\setlength{\tabcolsep}{10pt}
\begin{tabular}{lcc}
\toprule
 & \textbf{Generative answers} & \textbf{Reconstructions} \\
\midrule
Reference & Answer key & Original question \\
Avg. word length & 3.4 & 82.0 \\
\addlinespace[4pt]
\multicolumn{3}{l}{\textbf{Word overlap (\%)}} \\
\quad Judged as correct & 71.1 & 18.5 \\
\quad Judged as incorrect & 20.8 & 10.9 \\
\quad AUC [95\% CI] & 0.835 [0.826, 0.844] & 0.631 [0.616, 0.646] \\
\addlinespace[4pt]
\multicolumn{3}{l}{\textbf{Embedding similarity}} \\
\quad Judged as correct & 0.695 & 0.571 \\
\quad Judged as incorrect & 0.284 & 0.431 \\
\quad AUC [95\% CI] & 0.871 [0.862, 0.880] & 0.734 [0.719, 0.749] \\
\bottomrule
\end{tabular}
\end{table}

\subsection*{Expert evaluation}

We hired an independent expert, a licensed veterinarian (DVM) with training in physiology, pathology, pharmacology and clinical reasoning, to review 50 of the judge's verdicts on generative answers. The answers were drawn at random from MedQA-USMLE, MedMCQA and MedBullets: 25 that the judge said correct and 25 that it said incorrect, with at most five from any one model. For each answer, the expert saw the question, the answer key, the model's answer and the judge's verdict with its reason, rated whether the verdict and the reason were right, and could flag an answer key that looked wrong. The expert agreed with the judge on 49 of the 50 verdicts and disagreed with none. They agreed with all 25 rejections (100\%) and 24 of the 25 acceptances (96\%), rated the judge's reason sound in 48 of the 50 items (96\%), and flagged two answer keys. Supplementary Material S17 shows three annotated items (Figure~\ref{fig:supp-annotation}).

\section*{RELATED WORK}

\noindent\textbf{Medical question answering with language models.} General-purpose and domain-adapted language models have been evaluated on medical licensing and entrance examinations for several years, and reported accuracy on MedQA-USMLE \cite{app11146421}, MedMCQA \cite{pal2022medmcqa}, and PubMedQA \cite{jin2019pubmedqa} has risen steadily as models scaled and as prompting improved \cite{nori2023can,singhal2023large,maharjan2024openmedlm}. Chain-of-thought prompting \cite{wei2022chain} was a turning point: eliciting step-by-step solutions lifted these models from below the passing mark into the range of human examinees \cite{lievin2024can}, and subsequent adapted methods and frontier models pushed the reported numbers higher still \cite{wang2022self,yao2023tree}. There are works on gathering these benchmarks into the MultiMedQA suite. Flan-PaLM set the state of the art on every multiple-choice set in it, yet human evaluation of its answers revealed key gaps\cite{singhal2023large}. GPT-4 exceeded the USMLE passing score by more than 20 points without specialized prompting\cite{nori2023capabilities}. Open models have also been adapted to medicine, from MEDITRON, pretrained further on medical text\cite{chen2023meditron}, to MedGemma, whose 4B and 27B models we evaluate\cite{sellergren2025medgemma}. Across this line of work the measurement itself held in multiple domains both in multiple-choice and against the answer key so a higher number was read directly as greater medical ability. We adopt the same benchmarks, together with a recent board-preparation set, and the same protocol and ask what that number now certifies once models have mastered the format.

\noindent\textbf{Answer format and open-ended evaluation.} The multiple-choice format shapes how a model answers. Showing the question with its choices and asking for a letter, as in our Setting 1, lets the model compare the options directly\cite{robinson2023leveraging}. The format also has known artifacts. Models prefer particular option letters, so their answers can change when the options are reordered\cite{zheng2024large}, and they often beat a majority baseline when shown only the choices\cite{balepur2024artifacts}. Formats closer to clinical practice are harder. When MedQA cases are posed as sequential clinical encounters, diagnostic accuracy can drop below a tenth of the original\cite{schmidgall2024agentclinic}, and dialogues with simulated patients expose weaknesses in history-taking and diagnosis\cite{johri2025evaluation}. Other work argues for evaluation closer to clinical use\cite{yang2025beyond} or builds new interactive tasks in which the model must ask for missing information\cite{li2024mediq}. We instead withhold the choices from the same benchmark questions answered by the same models, so that the effect of the choices is measured within each model rather than across tasks.

\noindent\textbf{Contamination and memorization in large language models.} Because public benchmarks are widely exposed on the web, they are likely to enter the pretraining corpora \cite{ramos2025large,menta2025analyzing} of large models, and controlled studies show that models can memorize and generate training examples \cite{ishihara-2023-training}. Direct inference is generally impossible for closed models, whose training data is undisclosed, and the field has therefore turned to behavioral probes---prompting a model to complete, date, or reproduce a benchmark instance as indirect evidence of prior exposure \cite{golchin2024time,deng2024investigating,balepur-etal-2025-reverse}. Guided prompting, for example, gives the model the name of a dataset and the first part of an instance, and flags contamination when the completion matches the rest\cite{golchin2024time}. Other tests use the probability that the model assigns to a text, token by token\cite{shi2023detecting} or for a whole benchmark in its published and shuffled orders\cite{oren2023proving}, and so require access to those probabilities. In medicine, GPT-4 has been probed for memorization of examination content\cite{nori2023capabilities}. Our reconstruction setting belongs to this family: rather than asking whether a model has memorized an answer, it asks whether the model can reproduce the hidden question from its answer choices alone, which we treat as a proxy for LLM memorization of the question. It differs from reverse question answering\cite{balepur-etal-2025-reverse}, in which a model writes a question for a given answer: our target is one specific benchmark question, and the model sees that question's full set of choices. It is closer to testset slot guessing, which hides one wrong choice and asks the model to fill it in. GPT-4 recovered the hidden MMLU option exactly 57\% of the time this way\cite{deng2024investigating}. We instead withhold the entire question, a probe that needs only text output and thus applies to both open- and closed-weight models. Because models can sometimes infer the original question from the choices alone \cite{balepur2024artifacts}, we treat reconstruction only as a proxy for memorization. Unlike prior work that estimates contamination rates \cite{sainz2023nlp,xu2024benchmark}, we link each reconstruction to whether the same question was answered correctly, combining this probe with standard and open-ended evaluation. This complements benchmarks that avoid contamination by refreshing questions, such as LiveBench \cite{white2025livebench}.

\noindent\textbf{LLM-as-a-judge.} Scoring generated answers at scale has motivated the use of language models as judges \cite{zheng2023judging,gu2024survey}, which agree with human raters closely enough to serve as evaluators on many tasks \cite{szymanski2025limitations,williams2026human}. The approach has documented failure modes: judges are sensitive to answer length, position, and presentation, and can favour text generated by themselves or by similar models \cite{ye2025justice}. These biases matter when the judge grades open-ended clinical answers, where a lenient or drifting rubric could silently move headline numbers. We therefore fix a single rubric and use a single judge across all models and conditions, and validate the rubric against a more lenient variant so that the grading instrument is reported rather than assumed.

\noindent\textbf{Benchmark errors and ensembles.} Widely used test sets contain wrong labels. Across ten common machine-learning test sets, at least 3.3\% of labels are wrong on average, and a few percent more errors can reverse which of two models ranks higher\cite{northcutt2021pervasive}. An estimated 6.49\% of MMLU questions contain errors\cite{gema2024done}. Finding 8 flags the questions that every model misses for the same kind of review. Majority voting over repeated calls to one model raises accuracy as calls are added\cite{li2024more}, but accuracy can also rise and then fall, because extra calls help on easy questions and hurt on hard ones\cite{chen2024more}. Our committees instead vote across different models whose individual accuracies are known, and we weigh any gain against the added cost (Finding 7).

\section*{DISCUSSION}

We evaluated twelve current language models on four medical benchmarks under three settings: standard multiple choice, the same questions with the answer choices withheld, and reconstruction of a hidden question from its choices alone. Under multiple choice the strongest models exceed the human-expert baseline on MedQA-USMLE and PubMedQA and surpass the best 2023 models on every benchmark with earlier results \cite{lievin2024can}, continuing a trajectory read as steadily improving medical competence \cite{li2024mediq}.

Although these models score excellently on the benchmarks, at or above expert level, they cannot answer the same questions nearly as well once we change the setting. Withholding the answer choices and asking for the answer directly (Setting 2) cuts mean accuracy on MedQA-USMLE, MedMCQA and MedBullets from 82.3\% to 50.9\%, a drop that holds for every model and reorders the ranking. The effect is not confined to general-purpose systems: the two medically specialised MedGemma models lose more than any other model in the cohort, 40.0 and 37.6 points. Nothing about the medical content changed between the two settings: the questions, the models, and the decoding are identical, and the printed choices are removed. The most direct reading is that a substantial part of what the multiple-choice score measures is not medical knowledge but the ability to use the choices themselves, either by eliminating implausible ones or by recognising the answer once it is shown. This is consistent with evidence that answer choices leak information about the answer \cite{balepur2024artifacts,johri2025evaluation,yang2025beyond,kim2025fine}, and it means a score obtained with the choices visible should not be read as a measure of what a model could produce unaided.

Because the setting without answer choices is graded by an LLM judge, we also perform validation of the LLM judge itself. we shot that the judge's verdicts is correlated with the similarity between mean embeddings of original and generated answers. Additionally, an clinician expert agreed with 49 of 50 verdicts, with no wrong rejection. Where an answer repeats a choice word for word, the judge is exact: it accepted all 3,892 that repeat the correct choice and rejected all 482 that repeat a wrong one.

Another aspect we analyse is whether the models already know these benchmarks. Shown only the four answer choices and asked to reconstruct the question they belong to, the strongest models recover about 25\% of them, and reconstruction rates track multiple-choice accuracy closely. We then split each model's questions by whether it could reconstruct them. Models score 88.4\% on the questions they can reconstruct against 81.0\% on those they cannot, a gap of 7.4 percentage points, and once the answer choices are withheld the same gap widens to 14.5 points (62.9\% against 48.4\%). The comparison is made within benchmarks rather than across them, which rules out one dataset simply being easier than another. Two things temper the reading. Question difficulty is a plausible confound, since an easier question is easier both to answer and to reconstruct. The un-reconstructable pool is also the large majority of the questions, between 73.2\% and 90.7\% of them, and accuracy there stays close to each model's overall score. Prior familiarity therefore appears to contribute at the margin rather than to carry the score, but the effect is consistent across all twelve models and doubles, in percentage points, once the answer choices are removed, and we think it deserves more work than a single behavioural probe can settle.

Health information generally cannot leave an institution's control, and as a consequence, an AI system running on-premises inside a HIPAA-compliant boundary is often the only kind a clinical team can use. This constrains the selection to open-weight models that the institution can control \cite{magnini2025open}. Our analysis shows that most of the smaller models are nowhere near the best performing one, trailing the expert level by 8 to 21 percentage points. Local deployment is therefore close enough on multiple choice to be worth considering, but these models lose more than the frontier does when the answer choices are removed. Without the choices, the gap to their larger model counterparts widens from 10.1 to 13.4 points, and it widens in all six provider pairs, so multiple-choice scores make the smaller models look closer to the frontier than they are.

On cost analysis, accuracy per dollar strongly favors smaller models: one reaches within 1.5 percentage points of the best model at one-seventh the price. The smaller models nonetheless remain far below expert level, which makes the choice among smaller models matter more than the gap to the frontier. Ensembling does not close that gap. Majority voting gains at most 1.1 percentage points even when the committee is drawn from the six larger models, never helps among the smaller models, and multiplies inference cost. One caveat applies to all of these figures. They price hosted inference rather than self-hosting, which is the mode a HIPAA boundary actually requires, and the cost ratios rather than the absolute values are therefore what carry over.

Our results point to the evaluation, rather than the models, as the current limiting factor. A benchmark that can be solved by recognising the answer among four answer choices, and that models can partly reconstruct from those choices alone, is no longer as effective as it used to be. Almost half of all questions are now answered correctly by every model and fewer than four percent defeat all twelve models, and most of those few draw broad agreement on an answer the key rejects. We would not retire these benchmarks, but we would ask more of them: withhold the answer choices, allow more than one defensible answer, and report the reasoning with the same weight as the final choice.

These recommendations can be made concrete in three ways. First, medical benchmarks need a `refresh cycle'. An example benchmark for general tasks that performs such refreshes is LiveBench\cite{white2025livebench}: a private held-out split, new questions released on a dated schedule, and scores reported separately for questions released after a model's training cutoff. Second, reconstruction from the choices is cheap to run and can screen new questions before a benchmark release; a question that several models can reconstruct should be rewritten, paraphrased and validated by clinicians. Third, open-ended evaluation should mirror clinical practice, with cases that reveal findings in stages\cite{li2024mediq,schmidgall2024agentclinic}, clinician-written answer keys that accept alternative answers, and graders validated against clinicians\cite{johri2025evaluation}.

There is also room for work on deciding which model to trust on a given question or specialty, since the knowledge needed to answer these benchmarks is spread across models. Medical question answering is not solved. It is measured in a way that makes it look closer to solved than it is.

\section*{RESOURCE AVAILABILITY}

\subsection*{Lead contact}

Requests for further information and resources should be directed to and will be fulfilled by the lead contact, Mayank Kejriwal (kejriwal@isi.edu).

\subsection*{Materials availability}

This study did not generate new physical materials or reagents.

\subsection*{Data and code availability}

Every artifact described below --- model outputs, judge verdicts, derived data, and all original code --- is deposited in a single public, versioned archive at \url{https://doi.org/10.5281/zenodo.21736709} under a CC-BY-4.0 license, and is publicly available as of the date of this submission. The code is additionally mirrored for browsing at \url{https://github.com/navapatn/medical-qa-llm-evaluation}. The DOI above is the concept DOI and always resolves to the most recent version of the deposit.

\begin{itemize}
   \item \textbf{Source data.} This study generated no new primary data. All four benchmarks are publicly available from their original sources. The exact evaluation sets used here are the 1,273-question MedQA-USMLE test split, a fixed seed-0 1,000-question subset of the MedMCQA validation split (the MedMCQA test labels are not public), the 500-question PubMedQA set, and the 303-question MedBullets evaluation set. Datasets and outputs of each question set are deposited with the analysis artifacts so that the exact subsets can be reconstructed rather than approximated.

   \item \textbf{Code.} All original code will be deposited in the same repository and archived under the same DOI, and will be publicly available as of the date of publication under an open-source license. The release comprises the evaluation harness, the judge execution and finalization scripts, the statistical analysis pipeline that produces every reported test, and the generator for the interactive report. It also includes the full text of the prompt templates for all three conditions, the full text of both the canonical and the lenient judge rubrics, and the per-run configuration files, so that both the evaluation and the grading instrument can be reproduced exactly rather than reimplemented from description.

   \item \textbf{Model access and reproducibility.} All fourteen evaluated models and the judge model were accessed through commercial inference APIs rather than run locally, with the judge fixed to Claude Sonnet 5 at temperature 0. Provider-hosted models can be updated or withdrawn without notice, and as a consequence, re-querying the same model identifiers at a later date is not guaranteed to reproduce these generations. The deposited outputs, not live re-querying, therefore constitute the reproducible record of this study. The deposited code will regenerate every table, figure, and statistical test in this paper from those stored outputs without any API access.

   \item Any additional information required to reproduce the analyses reported in this paper is available from the lead contact upon request.
\end{itemize}

\section*{ACKNOWLEDGMENTS}

This research received no specific grant from any funding agency in the public, commercial, or non-profit organization.

\section*{AUTHOR CONTRIBUTIONS}

Conceptualization, MK and NN; methodology, NN; software, NN; investigation, MK; formal analysis, NN; writing--original draft, NN; writing--review \& editing, MK; supervision, MK.

\section*{DECLARATION OF INTERESTS}

The authors declare no competing interests.

\section*{DECLARATION OF GENERATIVE AI AND AI-ASSISTED TECHNOLOGIES}

During the preparation of this work, the author(s) used GRAIL AI latex editor to help on the writing. After using this tool, the author(s) reviewed and edited the content as needed and take(s) full responsibility for the content of the publication. Language models also served as research instruments, as described in Methods and the Supplementary Materials: Claude Sonnet 5 as the judge; and Claude Opus 5.5 to write codes for analysis work.

\newpage

\section*{SUPPLEMENTARY MATERIAL}

\setcounter{table}{0}
\renewcommand{\thetable}{S\arabic{table}}
\setcounter{figure}{0}
\renewcommand{\thefigure}{S\arabic{figure}}
\renewcommand{\theHfigure}{supp\arabic{figure}}
\renewcommand{\theHtable}{supp\arabic{table}}
\makeatletter
\let\suppOrigMakecaption\@makecaption
\renewcommand{\@makecaption}[2]{{\footnotesize\suppOrigMakecaption{#1}{#2}}}
\makeatother

This supplement gives the full numerical results underlying the figures in the main text (sections S1--S6), followed by the exact prompts and grading instruments used in the study and examples of model output (sections S7--S11), and the analyses that validate the judge, the reconstruction probe, the answer keys and the statistics (sections S12--S15). Section S16 reports two additional open-weight models of reported size, and Section S17 shows examples from the expert review of the judge.

\subsection*{S1. Model size across the three settings}

Table~\ref{tab:rq2scale} gives the within-provider size comparison behind Finding~2, extended to all three settings. The larger model leads on average in all three.

\begin{table}[H]
\centering
\scriptsize
\caption{Within-developer size comparisons across the three evaluation settings (difference = larger \mbox{$-$} smaller model, percentage points). Paired \mbox{$t$}-tests across the six families: multiple choice \mbox{$t(5)=3.37$}, \mbox{$p=.020$}; generative \mbox{$t(5)=4.41$}, \mbox{$p=.007$}; reconstruction \mbox{$t(5)=3.51$}, \mbox{$p=.017$}. Multiple choice covers all 3,076 questions, and the generative and reconstruction settings the 2,576 questions of MedQA-USMLE, MedMCQA and MedBullets. The last rows give the two added open-weight models of reported size (Supplementary Material~S16), which are not included in the tests; Qwen 3.5 35B ran with its reasoning mode off, against the thinking-mode Qwen 3.5 9B in multiple choice.}
\label{tab:rq2scale}
\begin{tabular}{llccc}
\toprule
\textbf{Developer} & \textbf{Model pair (smaller $\rightarrow$ larger)} & \textbf{MC} & \textbf{Gen} & \textbf{Recon} \\
\midrule
OpenAI & GPT-OSS 20B $\rightarrow$ GPT-5.6 Luna & +10.9 & +16.0 & +10.4 \\
DeepSeek & V4 Flash $\rightarrow$ V4 Pro & +0.0 & +1.1 & \mbox{$-0.1$} \\
Qwen & 3.5 9B $\rightarrow$ 3.7 Plus & +5.5 & +14.3 & +12.9 \\
Meta & Llama 3.1 8B $\rightarrow$ Llama 4 Maverick & +18.7 & +23.3 & +8.0 \\
Mistral & Ministral 8B $\rightarrow$ Mistral Large 3 & +10.4 & +15.9 & +9.8 \\
Google & MedGemma 4B \mbox{$\rightarrow$} MedGemma 27B & +6.6 & +9.7 & +2.4 \\
\midrule
\multicolumn{2}{l}{\textit{Mean paired difference}} & +8.7 & +13.4 & +7.2 \\
\midrule
\multicolumn{5}{l}{\textit{Added open-weight models of reported size (Supplementary Material S16)}}\\
OpenAI & GPT-OSS 20B \mbox{$\rightarrow$} GPT-OSS 120B & +6.1 & +6.9 & +6.0 \\
Qwen & 3.5 9B \mbox{$\rightarrow$} 3.5 35B & +1.4 & +8.4 & +4.6 \\
\multicolumn{2}{l}{\textit{Mean, these two pairs in place of the closed models}} & +7.2 & +10.9 & +5.1 \\
\bottomrule
\end{tabular}
\end{table}

\subsection*{S2. Accuracy with and without the answer choices}

Table~\ref{tab:rq4} gives the per-model, per-benchmark figures behind Finding~3 and Figure~\ref{fig:main2}.

\begin{table}[H]
\centering
\small
\caption{Multiple-choice (MC) and generative (Gen; answer choices withheld) accuracy (\%) on MedQA-USMLE, MedMCQA and MedBullets, which underpin the open-ended claim. ``Overall'' and \mbox{$\Delta$} are example-weighted over the 2,576 MedQA-USMLE, MedMCQA and MedBullets questions (MedBullets in its four-choice version), with \mbox{$\Delta=\text{Gen}-\text{MC}$} computed from unrounded values. PubMedQA is excluded because its \emph{yes}, \emph{no}, or \emph{maybe} answers make the generative setting non-open-ended. The last rows give the two added open-weight models of reported size (Supplementary Material~S16), which are not included in the means or tests.}
\label{tab:rq4}
\resizebox{0.75\linewidth}{!}{%
\begin{tabular}{lccccccccc}
\toprule
\multirow{2}{*}{\textbf{Model}} & \multicolumn{2}{c}{\textbf{MedQA-USMLE}} & \multicolumn{2}{c}{\textbf{MedMCQA}} & \multicolumn{2}{c}{\textbf{MedBullets}} & \multicolumn{2}{c}{\textbf{Overall}} & \multirow{2}{*}{\mbox{$\Delta$}} \\
\cmidrule(lr){2-3}\cmidrule(lr){4-5}\cmidrule(lr){6-7}\cmidrule(lr){8-9}
 & MC & Gen & MC & Gen & MC & Gen & MC & Gen & \\
\midrule
GPT-5.6 Luna & 94.7 & 76.5 & 83.8 & 52.1 & 89.1 & 67.3 & 89.8 & 65.9 & \mbox{$-23.9$} \\
Qwen 3.7 Plus & 94.7 & 70.2 & 83.8 & 48.5 & 87.8 & 61.4 & 89.6 & 60.7 & \mbox{$-28.9$} \\
DeepSeek V4 Pro & 94.4 & 69.1 & 82.4 & 48.1 & 87.5 & 61.1 & 88.9 & 60.0 & \mbox{$-28.9$} \\
DeepSeek V4 Flash & 94.0 & 68.9 & 81.8 & 46.3 & 88.4 & 58.7 & 88.6 & 58.9 & \mbox{$-29.7$} \\
Llama 4 Maverick & 89.9 & 60.8 & 79.4 & 44.7 & 79.2 & 51.2 & 84.5 & 53.4 & \mbox{$-31.2$} \\
Mistral Large 3 & 88.8 & 65.6 & 80.8 & 47.6 & 82.2 & 52.5 & 84.9 & 57.1 & \mbox{$-27.8$} \\
Qwen 3.5 9B & 91.9 & 55.2 & 74.1 & 36.8 & 77.2 & 41.9 & 83.3 & 46.5 & \mbox{$-36.8$} \\
GPT-OSS 20B & 85.3 & 62.0 & 68.8 & 34.5 & 72.3 & 50.2 & 77.4 & 49.9 & \mbox{$-27.4$} \\
Ministral 8B & 77.4 & 49.1 & 69.3 & 32.5 & 63.0 & 37.0 & 72.6 & 41.2 & \mbox{$-31.3$} \\
Llama 3.1 8B & 67.5 & 35.1 & 58.4 & 26.0 & 57.4 & 22.8 & 62.8 & 30.1 & \mbox{$-32.6$} \\
MedGemma 27B & 92.0 & 56.1 & 79.7 & 38.3 & 82.5 & 50.2 & 86.1 & 48.5 & \mbox{$-37.6$} \\
MedGemma 4B & 85.5 & 43.1 & 73.0 & 32.6 & 69.3 & 40.9 & 78.8 & 38.8 & \mbox{$-40.0$} \\
\midrule
\textit{Mean} & 88.0 & 59.3 & 76.3 & 40.7 & 78.0 & 49.6 & 82.3 & 50.9 & \mbox{$-31.3$} \\
\midrule
\multicolumn{10}{l}{\textit{Added open-weight models of reported size (Supplementary Material S16)}}\\
GPT-OSS 120B & 90.6 & 68.2 & 75.5 & 42.4 & 84.2 & 56.8 & 84.0 & 56.8 & \mbox{$-27.1$} \\
Qwen 3.5 35B & 91.9 & 65.3 & 77.3 & 42.0 & 81.2 & 53.1 & 85.0 & 54.8 & \mbox{$-30.1$} \\
\bottomrule
\end{tabular}}
\end{table}

\subsection*{S3. How often models reconstruct the hidden question}

Table~\ref{tab:rq3} gives the reconstruction rates behind Finding~4 and Figure~\ref{fig:main3}a for each benchmark, with a 95\% confidence interval on the pooled rate.

\begin{table}[H]
\centering
\scriptsize
\caption{Question-reconstruction fidelity (\%). Reconstruction is defined for MedQA-USMLE, MedMCQA and MedBullets (four-choice version). PubMedQA is excluded because its \emph{yes}, \emph{no}, or \emph{maybe} choices are identical across questions and carry nothing to reconstruct. Brackets give 95\% confidence intervals on the overall rate, which pools MedQA-USMLE, MedMCQA and MedBullets. The last rows give the two added open-weight models of reported size (Supplementary Material~S16), which are not included in the means or tests.}
\label{tab:rq3}
\begin{tabular}{lccccc}
\toprule
\textbf{Model} & \textbf{MedQA-USMLE} & \textbf{MedMCQA} & \textbf{MedBullets} & \textbf{Overall} & \textbf{95\% CI} \\
\midrule
Qwen 3.7 Plus & 30.27 & 21.50 & 29.37 & 26.76 & [25.1, 28.5] \\
GPT-5.6 Luna & 29.12 & 21.10 & 28.05 & 25.88 & [24.2, 27.6] \\
DeepSeek V4 Flash & 24.03 & 17.90 & 20.13 & 21.18 & [19.6, 22.8] \\
Mistral Large 3 & 24.37 & 15.20 & 25.08 & 20.89 & [19.4, 22.5] \\
DeepSeek V4 Pro & 22.24 & 17.70 & 27.39 & 21.08 & [19.5, 22.7] \\
Llama 4 Maverick & 18.10 & 14.50 & 22.85 & 17.26 & [15.8, 18.8] \\
GPT-OSS 20B & 18.30 & 11.91 & 15.51 & 15.49 & [14.1, 16.9] \\
Qwen 3.5 9B & 15.38 & 11.50 & 14.85 & 13.81 & [12.5, 15.2] \\
Ministral 8B & 12.35 & 9.02 & 12.54 & 11.08 & [9.9, 12.4] \\
Llama 3.1 8B & 9.28 & 8.90 & 10.56 & 9.29 & [8.2, 10.5] \\
MedGemma 27B & 19.34 & 16.70 & 15.84 & 17.90 & [16.5, 19.4] \\
MedGemma 4B & 16.99 & 12.60 & 19.14 & 15.54 & [14.2, 17.0] \\
\midrule
\multicolumn{6}{l}{\textit{Added open-weight models of reported size (Supplementary Material S16)}}\\
GPT-OSS 120B & 24.63 & 17.00 & 23.43 & 21.52 & [20.0, 23.2] \\
Qwen 3.5 35B & 22.25 & 12.90 & 20.46 & 18.41 & [17.0, 20.0] \\
\bottomrule
\end{tabular}
\end{table}

\subsection*{S4. Reconstruction split by whether the answer was correct}

Table~\ref{tab:rq9} gives, for each model, the reconstruction rates behind the 7.8-point difference in Finding~4, split by whether the model answered the question correctly.

\begin{table}[H]
\centering
\scriptsize
\caption{Question-reconstruction fidelity (\%) split by whether the model answered the multiple-choice question correctly. Reconstruction is defined for MedQA-USMLE, MedMCQA and MedBullets, which are pooled. The last rows give the two added open-weight models of reported size (Supplementary Material~S16), which are not included in the means or tests.}
\label{tab:rq9}
\begin{tabular}{lccc}
\toprule
\textbf{Model} & \textbf{Reconstruction when correct} & \textbf{Reconstruction when wrong} & \textbf{Difference} \\
\midrule
Qwen 3.7 Plus & 28.1 & 15.4 & +12.7 \\
GPT-5.6 Luna & 26.9 & 17.1 & +9.8 \\
DeepSeek V4 Flash & 22.4 & 12.0 & +10.4 \\
Mistral Large 3 & 22.1 & 14.1 & +8.0 \\
DeepSeek V4 Pro & 21.9 & 14.7 & +7.1 \\
GPT-OSS 20B & 17.5 & 8.8 & +8.7 \\
Llama 4 Maverick & 18.1 & 12.6 & +5.6 \\
Qwen 3.5 9B & 14.7 & 9.3 & +5.4 \\
Ministral 8B & 12.2 & 8.2 & +4.0 \\
Llama 3.1 8B & 11.0 & 6.4 & +4.6 \\
MedGemma 27B & 19.3 & 9.2 & +10.1 \\
MedGemma 4B & 17.1 & 9.7 & +7.4 \\
\midrule
\multicolumn{4}{l}{\textit{Paired \mbox{$t$}-test (\mbox{$n=12$}): mean \mbox{$+7.8$}, \mbox{$t(11)=10.25$}, \mbox{$p<.0001$}}}\\
\midrule
\multicolumn{4}{l}{\textit{Added open-weight models of reported size (Supplementary Material S16)}}\\
GPT-OSS 120B & 23.1 & 13.3 & +9.7 \\
Qwen 3.5 35B & 19.5 & 12.1 & +7.4 \\
\bottomrule
\end{tabular}
\end{table}

\subsection*{S5. Accuracy on reconstructable and un-reconstructable questions}

Table~\ref{tab:rq5} gives the pool split behind Finding~5 and the accuracies plotted in Figure~\ref{fig:main4}.

\begin{table}[H]
\centering
\small
\caption{Performance on each model's un-reconstructable pool: the MedQA-USMLE, MedMCQA and MedBullets questions it could not reproduce from the answer choices. Multiple-choice (MC) and generative (Gen) accuracy (\%) are given separately for the un-reconstructable (un-rec.) and reconstructable (rec.) pools. The last rows give the two added open-weight models of reported size (Supplementary Material~S16), which are not included in the means or tests.}
\label{tab:rq5}
\resizebox{0.75\linewidth}{!}{%
\begin{tabular}{lccccccc}
\toprule
\multirow{2}{*}{\textbf{Model}} & \textbf{Un-rec.} & \textbf{\% of} & \multicolumn{3}{c}{\textbf{Multiple choice}} & \multicolumn{2}{c}{\textbf{Generative}} \\
\cmidrule(lr){4-6}\cmidrule(lr){7-8}
 & \textbf{(n)} & \textbf{set} & un-rec. & rec. & diff. & un-rec. & rec. \\
\midrule
GPT-5.6 Luna & 1,905 & 74.1 & 88.6 & 93.2 & +4.7 & 63.1 & 74.0 \\
Qwen 3.7 Plus & 1,886 & 73.2 & 88.0 & 94.0 & +6.0 & 58.1 & 68.1 \\
DeepSeek V4 Pro & 2,029 & 78.9 & 88.0 & 92.3 & +4.2 & 57.1 & 70.7 \\
DeepSeek V4 Flash & 2,024 & 78.8 & 87.3 & 93.6 & +6.3 & 56.7 & 67.3 \\
Mistral Large 3 & 2,037 & 79.1 & 83.6 & 89.8 & +6.2 & 53.3 & 71.6 \\
Llama 4 Maverick & 2,129 & 82.7 & 83.7 & 88.7 & +5.1 & 51.4 & 63.1 \\
GPT-OSS 20B & 2,172 & 84.5 & 75.6 & 87.2 & +11.6 & 46.9 & 66.8 \\
Qwen 3.5 9B & 2,216 & 86.2 & 82.4 & 88.7 & +6.3 & 44.4 & 58.9 \\
Ministral 8B & 2,287 & 88.9 & 71.6 & 79.6 & +8.0 & 39.3 & 56.8 \\
Llama 3.1 8B & 2,335 & 90.7 & 61.6 & 74.5 & +12.9 & 28.8 & 43.1 \\
MedGemma 27B & 2,114 & 82.1 & 84.6 & 92.8 & +8.2 & 45.5 & 62.3 \\
MedGemma 4B & 2,174 & 84.5 & 77.3 & 86.8 & +9.5 & 36.2 & 52.5 \\
\midrule
\multicolumn{8}{l}{\textit{Paired test, rec.\ vs.\ un-rec.\ (multiple choice): mean \mbox{$+7.4$}, \mbox{$t(11)=9.41$}, \mbox{$p<.0001$}}}\\
\multicolumn{8}{l}{\textit{Paired test, rec.\ vs.\ un-rec.\ (generative): mean \mbox{$+14.5$}, \mbox{$t(11)=15.33$}, \mbox{$p<.0001$}}}\\
\midrule
\multicolumn{8}{l}{\textit{Added open-weight models of reported size (Supplementary Material S16)}}\\
GPT-OSS 120B & 2,020 & 78.5 & 82.3 & 90.1 & +7.7 & 54.0 & 67.1 \\
Qwen 3.5 35B & 2,101 & 81.6 & 83.8 & 90.1 & +6.3 & 51.8 & 68.4 \\
\bottomrule
\end{tabular}}
\end{table}

\subsection*{S6. Ensemble accuracy by pool}

Table~\ref{tab:ensemble} gives, by benchmark, the vote of all members of each pool, with the best single model of the pool for comparison; for the three pools of Figure~\ref{fig:ensemble}, this is the last point in each panel.

\begin{table}[H]
\centering
\small
\caption{Majority-vote accuracy (\%) against the best single model, for the full twelve-model pool and for the pools built from the larger and the smaller model of each developer family. Votes are cast on answer letters; a tie is broken by the highest-ranked member voting for one of the tied letters. Overall pools all 3,076 questions of the four benchmarks. Single-model accuracy for all twelve models is in Table~\ref{tab:rq1}. The last rows give the vote of all fourteen models and of the six larger models with GPT-OSS 120B and Qwen 3.5 35B in place of GPT-5.6 Luna and Qwen 3.7 Plus (Supplementary Material~S16); the best single model of the fourteen is still Qwen 3.7 Plus.}
\label{tab:ensemble}
\resizebox{0.75\linewidth}{!}{%
\begin{tabular}{lccccc}
\toprule
\textbf{Approach} & \textbf{MedQA-USMLE} & \textbf{MedMCQA} & \textbf{PubMedQA} & \textbf{MedBullets} & \textbf{Overall} \\
\midrule
\multicolumn{6}{l}{\textit{Full twelve-model pool and six-larger-model pool}}\\
Best single model in both pools (Qwen 3.7 Plus) & 94.7 & 83.8 & 81.6 & 87.8 & 88.3 \\
Majority vote (12 models) & 93.9 & 83.4 & 82.2 & 88.1 & 88.0 \\
Larger-6 majority vote & 95.1 & 85.3 & 80.8 & 89.1 & 89.0 \\
\midrule
\multicolumn{6}{l}{\textit{Six-smaller-model pool (smaller model per developer)}}\\
Best single smaller model (DeepSeek V4 Flash) & 94.0 & 81.8 & 77.8 & 88.4 & 86.8 \\
Smaller-6 majority vote & 92.6 & 78.3 & 80.0 & 84.5 & 85.1 \\
\midrule
\multicolumn{6}{l}{\textit{Pools with the added open-weight models (Supplementary Material S16)}}\\
Majority vote (14 models) & 94.2 & 83.2 & 82.0 & 88.1 & 88.0 \\
Larger, reported size: best (DeepSeek V4 Pro) & 94.4 & 82.4 & 76.0 & 87.5 & 86.8 \\
Larger, reported size: majority vote & 93.9 & 83.4 & 79.6 & 87.1 & 87.5 \\
\bottomrule
\end{tabular}}
\end{table}

\subsection*{S7. Prompt templates}

\begin{figure}[H]
\centering
\includegraphics[width=0.60\linewidth]{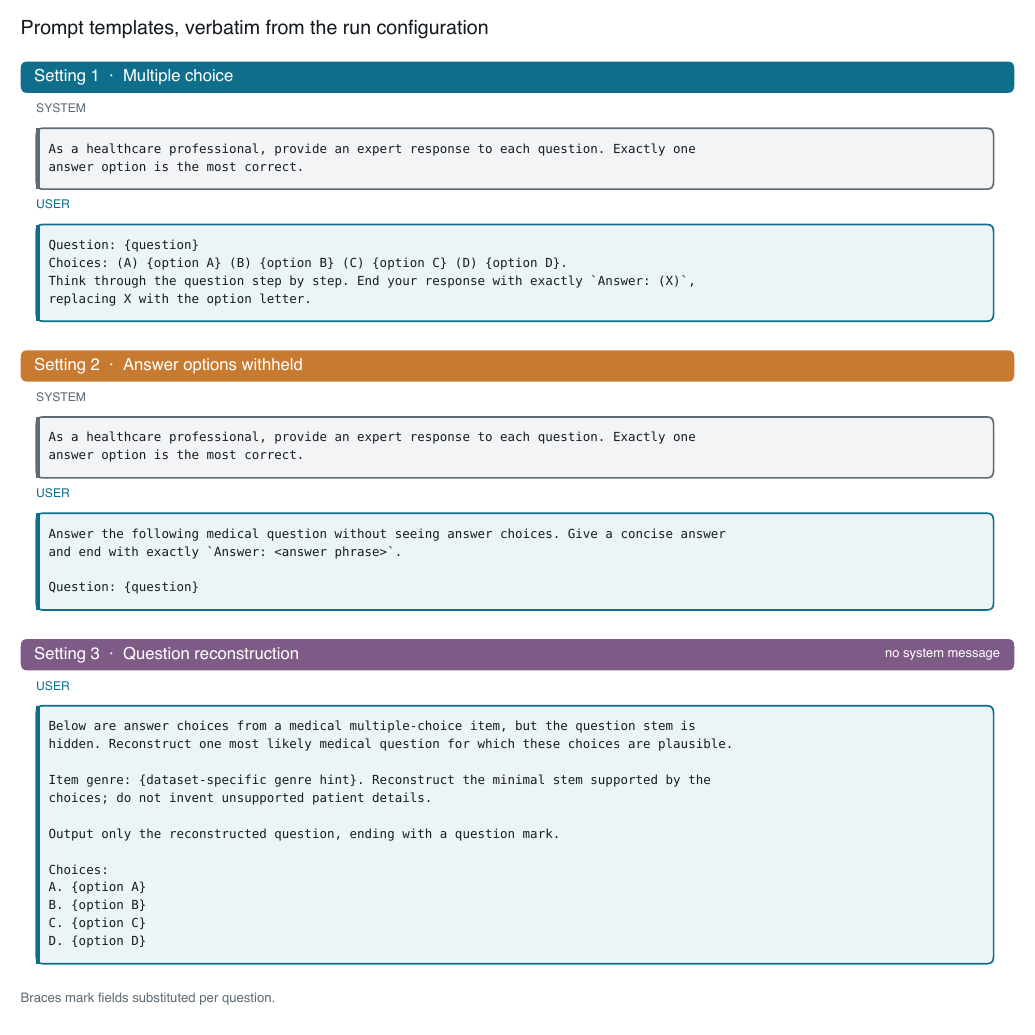}
\caption{The prompt issued in each of the three settings. Braces mark fields substituted per question; role labels indicate the message sequence sent to the provider.}
\label{fig:supp-prompts}
\end{figure}

Figure~\ref{fig:supp-prompts} reproduces the prompt for each of the three settings as it was issued to every model, with the per-question fields marked in braces. One difference between the settings bears on the comparisons in the main text: the genre hint in Setting 3 varies by dataset, telling the model whether the choices came from a licensing examination or an entrance examination, without naming the topic.

\subsection*{S8. Judge instruments}

\begin{figure}[H]
\centering
\includegraphics[width=0.72\linewidth]{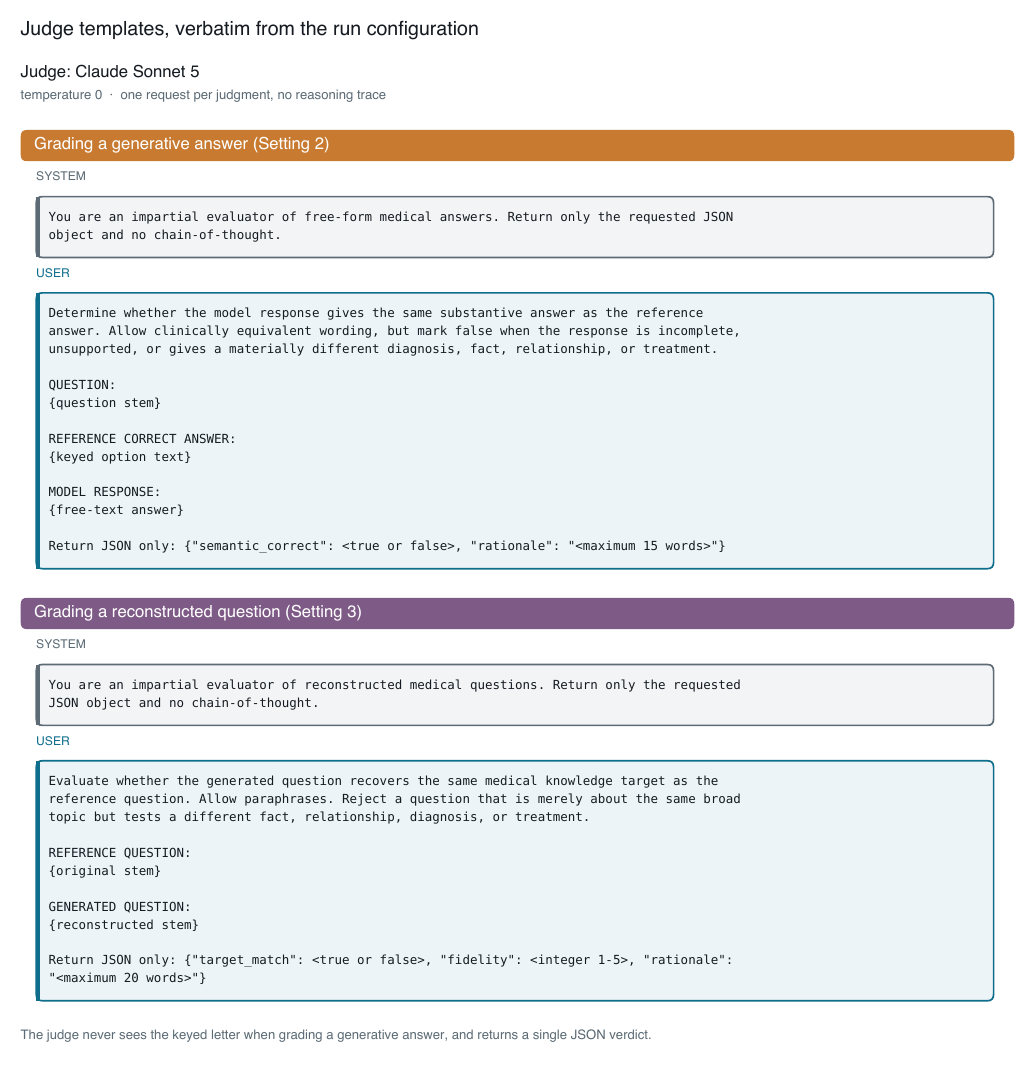}
\caption{The two grading prompts used by the judge. Braces mark fields substituted per judgment.}
\label{fig:supp-judge}
\end{figure}

Figure~\ref{fig:supp-judge} gives both grading prompts. Each judgment is one request: the judge receives the rubric, the question, the reference answer, and the model's output, and returns a JSON verdict with a short rationale and no reasoning trace. When grading a generative answer the judge is shown the keyed choice text but never the keyed letter, which prevents it from scoring by position. Total judge inference cost was approximately \$105 across all models and both settings.

\subsection*{S9. An example of model reasoning}

\begin{figure}[H]
\centering
\includegraphics[width=0.72\linewidth]{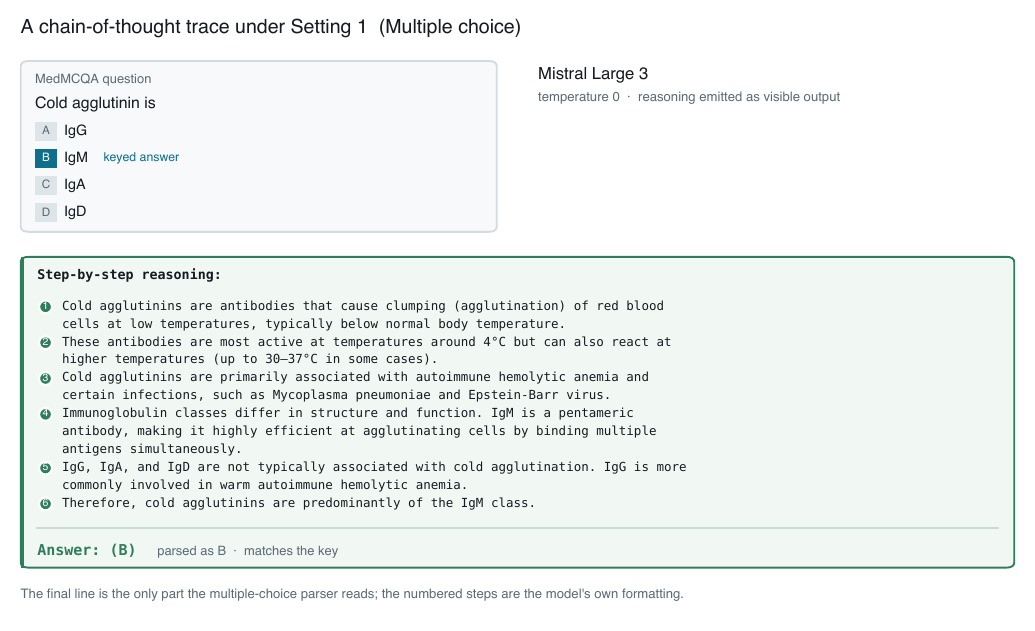}
\caption{A complete model output under Setting 1, with the question that produced it. The numbered steps are the model's own formatting.}
\label{fig:supp-reasoning}
\end{figure}

Figure~\ref{fig:supp-reasoning} shows one complete model output under Setting 1, chosen because it is short enough to reproduce in full. The model works through the question in numbered steps before committing to a letter. Models that have no separate reasoning channel emit their chain of thought as visible output, and as a consequence, their deliberation is billed as completion tokens.

\subsection*{S10. Correct and incorrect reasoning side by side}

\begin{figure}[H]
\centering
\includegraphics[width=0.72\linewidth]{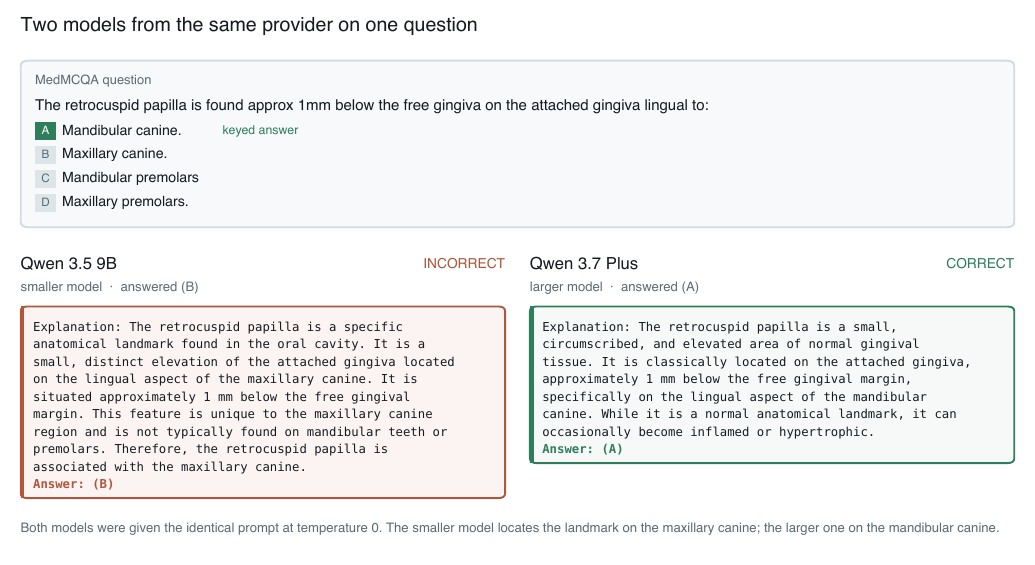}
\caption{The smaller and larger model from one provider on the same question, with the keyed answer marked. Both outputs are reproduced in full.}
\label{fig:supp-paired}
\end{figure}

Figure~\ref{fig:supp-paired} places the two Qwen models on a single MedMCQA question, one from each size tier, under the identical prompt. Both produce fluent, confidently worded explanations of the same length and form. They differ on one anatomical fact, and that single divergence decides the answer. The failure is not a breakdown in reasoning structure but a wrong premise carried confidently through to a conclusion, which is the typical shape of an incorrect answer in this cohort and the reason surface fluency is a poor guide to correctness.

\subsection*{S11. Questions that every model answered incorrectly}

\begin{figure}[H]
\centering
\includegraphics[width=0.72\linewidth]{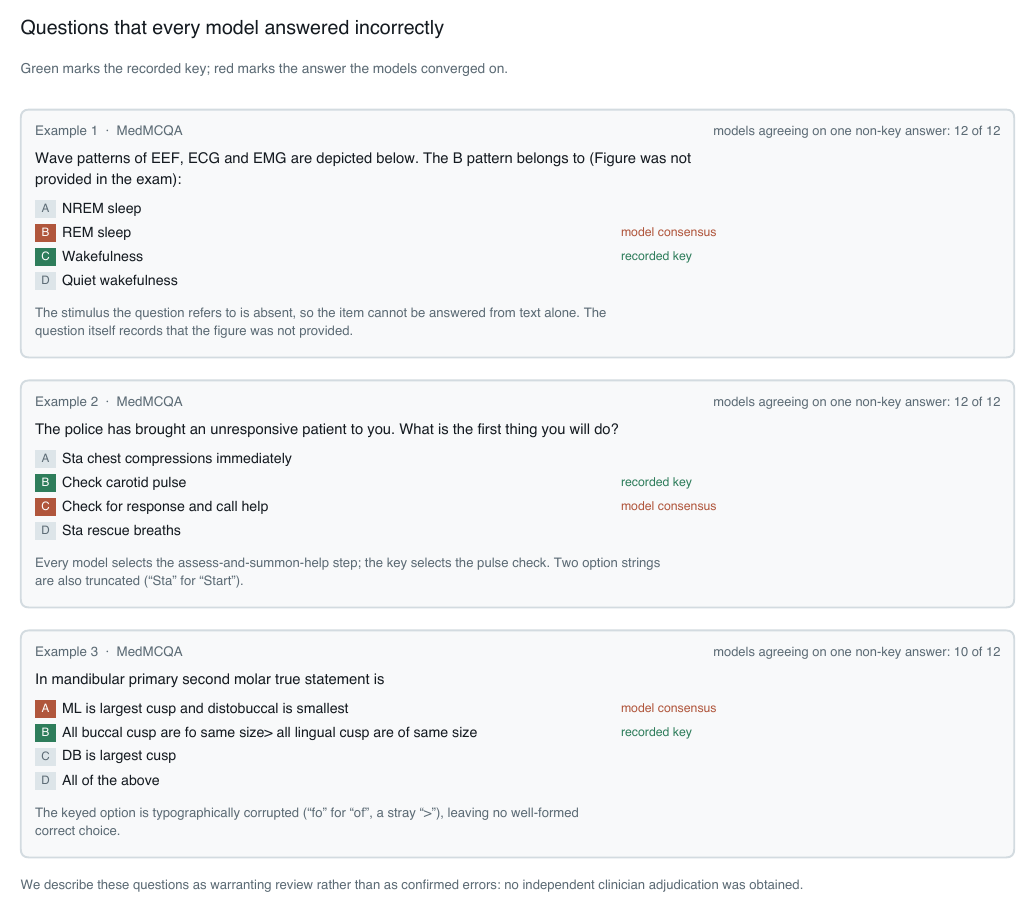}
\caption{Three questions answered incorrectly by all twelve models. The recorded key and the answer the models converged on are both marked; the note gives the pattern each question illustrates.}
\label{fig:supp-hard}
\end{figure}

Figure~\ref{fig:supp-hard} reproduces three of the 107 questions missed by all twelve models, one for each pattern described in Finding~8. In the first, the question refers to a figure that the dataset does not contain, and records as much in its own text, which leaves the question unanswerable from the text alone. In the second, every model selects the assess-and-summon-help step while the recorded key selects the pulse check, and two choice strings are truncated. In the third, the keyed choice is typographically corrupted, leaving no well-formed correct choice. In all three the models agree with one another and disagree with the key, which is the pattern that distinguishes a possibly defective question from a genuinely difficult one. We describe these questions as warranting review rather than as confirmed errors, since no independent clinician adjudication was obtained, and we release the full set of 107 with the analysis artifacts, identifying the five MedBullets questions by number only, since their text is not ours to redistribute.

\subsection*{S12. Agreement between the judge and text-similarity measures}

Table~\ref{tab:judge-nlp} adds chance levels, the range of AUCs across models and same-question differences to Table~\ref{tab:judge-similarity}. Word overlap is the share of the reference's summed inverse-document-frequency weight that the output recovers, with IDF computed over the 2,576 questions (for reconstructions) or the 2,576 answer keys (for answers). Embedding similarity uses all-MiniLM-L6-v2 with mean pooling and a 256-token limit, which truncates 229 of the longest questions. For answers, the embedding compares the phrase after the final ``Answer:'' with the key; word overlap uses the whole answer. Chance scores the same output against a random other question of the same benchmark. Outputs without a valid judge verdict (44 reconstructions, 40 answers) are left out, as is one question whose key has no words. The reconstruction AUCs are similar on each benchmark: word overlap 0.62 on MedQA-USMLE, 0.75 on MedMCQA and 0.63 on MedBullets; embedding similarity 0.70, 0.79 and 0.66. Split instead by whether the model answered the question correctly, reconstruction overlap barely moves (12.2\% against 12.5\%, AUC 0.50). Of the 15 reconstructions that share a run of 13 or more words with their original question, 2 were accepted by the judge; pairing with a random question gives 3 such runs by chance. No generative answer shares such a run with its key, but only 56 keys are that long.

\begin{table}[H]
\centering
\scriptsize
\caption{Additional results for Table~\ref{tab:judge-similarity}, over all outputs on the 2,576 MedQA-USMLE, MedMCQA and MedBullets questions. Chance: mean similarity when the same outputs are scored against a random other question of the same benchmark (word overlap in percent, embedding similarity as cosine). AUC by model: lowest and highest AUC across the twelve models. Same question: accepted minus rejected among different models answering the same question, with a 95\% interval from resampling questions.}
\label{tab:judge-nlp}
\begin{tabular}{llccc}
\toprule
\textbf{Output} & \textbf{Measure} & \textbf{Chance} & \textbf{AUC by model} & \textbf{Same question [95\% CI]} \\
\midrule
Answers & Word overlap & 3.9 & .79--.88 & 41.6 [39.8, 43.3] \\
 & Embedding & 0.100 & .84--.88 & 0.345 [0.333, 0.357] \\
\addlinespace[3pt]
Reconstructions & Word overlap & 5.6 & .59--.66 & 5.8 [5.2, 6.5] \\
 & Embedding & 0.209 & .70--.76 & 0.085 [0.079, 0.091] \\
\bottomrule
\end{tabular}
\end{table}

\subsection*{S13. Controls for the reconstruction probe }

\emph{Judge calibration and negative control.} We tested whether the Setting 3 judge accepts a reconstruction only when it asks the same thing as the original. We used pairs whose answer is known in advance. For a subsample of 120 benchmark questions, a model outside the cohort wrote a full paraphrase (a positive pair), a short restatement of the same target and a short question on a different fact. We also paired 300 accepted reconstructions (25 per model) with the most similar other question, and 156 with a random answers (injected negative pairs). Independent raters (Claude Opus 5.5, blind to the verdict) labelled the short and most-similar pairs. The judge accepted 119 of the 120 positive pairs and none of the 156 negative pairs, and agreed with the raters on 93.3\% of the 519 pairs they did not call borderline (\mbox{$\kappa=0.81$}; sensitivity 84\%, specificity 96\%). Conclusion: the judge is accurate and favor towards rejection.

\emph{Inspection.} We tested whether accepted reconstructions contain details that only memory could supply. We used a subsample of 120 accepted reconstructions, ten per model. Two raters (Claude Opus 5.5, run independently) read each one with the choices the model saw and the original question, and agreed on 97.5\% of items (\mbox{$\kappa=0.86$}). Both rated 107 as inferable from the choices and standard textbook associations and 10 as not testing the original target; neither found one that reproduced a distinctive detail of the original. Conclusion: accepted reconstructions show no sign of recall. Because the Setting 3 prompt asks for the minimal stem and no invented patient details, this check is weaker than the next one.

\emph{Membership inference.} We tested whether models reproduce a question's exact wording when told which dataset it comes from. Probability-based tests such as Min-K\% Prob\cite{shi2023detecting} need token probabilities, which the closed models and our API endpoints do not expose, so we used guided completion\cite{golchin2024time}. For a subsample of 180 questions of at least 30 words (70 MedQA-USMLE, 70 MedBullets, 40 MedMCQA), each model received the first half and was asked to finish it exactly as it appears in the named dataset (guided) or simply as a medical examination question (general), with reasoning off where the provider allows. We scored each completion against the true second half with ROUGE-L. Mean ROUGE-L was 22.4 guided against 22.3 general; six models scored significantly higher under the guided prompt and four significantly lower. Near-verbatim completions (ROUGE-L of at least 0.75, or a shared run of 13 words) were rare under both prompts, 10 of 2,160 guided and 4 of 2,160 general, and the guided gain did not track the reconstruction rate across models (\mbox{$r=0.03$}). Conclusion: naming the dataset does not help models reproduce question text, and we find no sign of verbatim memorization.


\subsection*{S14. Audit of answer keys}

An auditor (Claude Opus 5.5) judged the key of each of 612 questions: the 445 on which the four leading models disagree, the 107 missed by all twelve, and 60 drawn at random from the questions all twelve solve. The three groups were shuffled together and the auditor saw neither the group nor any model's answer, only the question, the choices and the key. It first chose its own answer, then rated the key as correct, correct but with an equally defensible alternative, ambiguous or flawed, wrong, or outdated. A second reviewer, instructed to defend the published key, re-examined every flag; a flag stands only if the reviewer also rejected the key. Of 136 first-pass flags, 125 stood. Removing all 125 flagged questions raises every model's accuracy by two to three points and leaves the order of the twelve models unchanged, except that DeepSeek V4 Flash (89.87\%) moves ahead of DeepSeek V4 Pro (89.56\%), with which it was tied at 86.83\%. Among the 94 unsolved questions where eight or more models agreed on a non-key answer, the flag stood in 59, and in 45 the reviewer's own answer was the models' answer. The MedBullets release also contains ten pairs of questions with identical stems, four of them with identical choices; we keep them as released. The audit is model-based. The flagged questions, with both rationales, are released for clinician review.

\begin{table}[H]
\centering
\small
\caption{Audit of answer keys. First pass: questions the auditor flagged. Kept: flags that stood after the second review, split by type.}
\label{tab:key-audit}
\resizebox{0.75\linewidth}{!}{%
\begin{tabular}{lccccccc}
\toprule
\textbf{Questions} & \textbf{n} & \textbf{First pass} & \textbf{Kept} & \textbf{Wrong} & \textbf{Outdated} & \textbf{Alternative} & \textbf{Ambiguous} \\
\midrule
Leading models disagree & 445 & 68 & 60 & 4 & 2 & 15 & 39 \\
Missed by all twelve & 107 & 68 & 65 & 18 & 1 & 22 & 24 \\
Solved by all twelve & 60 & 0 & 0 & 0 & 0 & 0 & 0 \\
\bottomrule
\end{tabular}}
\end{table}

\subsection*{S15. Multiplicity and question-level clustering}

Every model answers the same questions, so outcomes are clustered by question as well as by model, and the paired tests in the text treat models as the unit. We recomputed all 33 inferential results of the twelve-model analyses from the per-question data; every statistic and \mbox{$p$}-value matches. Table~\ref{tab:multiplicity} lists the 13 confirmatory tests with their Holm-\cite{holm1979simple} and Benjamini--Hochberg-adjusted\cite{benjamini1995controlling} \mbox{$p$}-values and a 95\% interval from a cluster bootstrap that resamples questions within each benchmark and keeps all twelve models' outcomes for each question (10,000 draws). At a 5\% false-discovery rate every result reported as significant remains so. Under the stricter Holm correction, the three comparisons that rest on only six provider pairs have adjusted \mbox{$p=.068$}. For the size comparisons, question sampling adds little uncertainty, which comes mostly from having only six provider pairs. For the reconstruction analyses, question sampling adds about as much uncertainty as the variation between models, and the \mbox{$p$}-values stay below .0001. Logistic GEE models with question clusters, model fixed effects and robust standard errors\cite{liang1986longitudinal} give the same conclusions. Two readings need care. The widening of the size gap without choices, and the doubling of the reconstruction gap, hold in percentage points but not on the odds scale, because multiple-choice accuracy is close to its ceiling. And the association between reconstruction and accuracy runs between questions: conditioning on the question with a conditional logistic regression removes it (odds ratio 1.07, \mbox{$p=.43$}).

\begin{table}[H]
\centering
\scriptsize
\caption{The 13 confirmatory tests with multiplicity-adjusted \mbox{$p$}-values and question-clustered 95\% intervals for the statistic reported in the text (points, or \mbox{$r$}). F: finding.}
\label{tab:multiplicity}
\begin{tabular}{lcccc}
\toprule
\textbf{Claim} & \textbf{\mbox{$p$}} & \textbf{Holm} & \textbf{BH} & \textbf{Clustered 95\% CI} \\
\midrule
Larger vs smaller, multiple choice (F2) & .020 & .068 & .024 & [8.0, 9.3] \\
Larger vs smaller, MedBullets (F2) & .010 & .049 & .014 & [11.0, 15.9] \\
Drop without choices (F3) & $<$.0001 & $<$.0001 & $<$.0001 & [-32.6, -30.1] \\
Reconstruction vs accuracy, $r$ (F4) & .0002 & .002 & .0005 & [0.83, 0.90] \\
Reconstruction higher if answered (F4) & $<$.0001 & $<$.0001 & $<$.0001 & [6.0, 9.6] \\
Accuracy higher if reconstructed, MC (F5) & $<$.0001 & $<$.0001 & $<$.0001 & [5.8, 9.0] \\
Accuracy higher if reconstructed, generative (F5) & $<$.0001 & $<$.0001 & $<$.0001 & [12.1, 16.9] \\
Unique wins less reconstructable (F5) & .001 & .009 & .002 & [-8.6, -4.4] \\
Twelve-model vote vs best model (F7) & .474 & .474 & .474 & [-1.1, 0.5] \\
Unanimous misses by benchmark (F8) & $<$.0001 & $<$.0001 & $<$.0001 & -- \\
Size gap wider without choices (Discussion) & .025 & .068 & .027 & [2.2, 4.4] \\
Larger vs smaller, generative (S1) & .007 & .042 & .011 & [12.5, 14.3] \\
Larger vs smaller, reconstruction (S1) & .017 & .068 & .022 & [6.5, 8.0] \\
\bottomrule
\end{tabular}
\end{table}

\subsection*{S16. Open-weight models of reported size for OpenAI and Qwen}

\begin{figure}[H]
\centering
\includegraphics[width=0.71\linewidth]{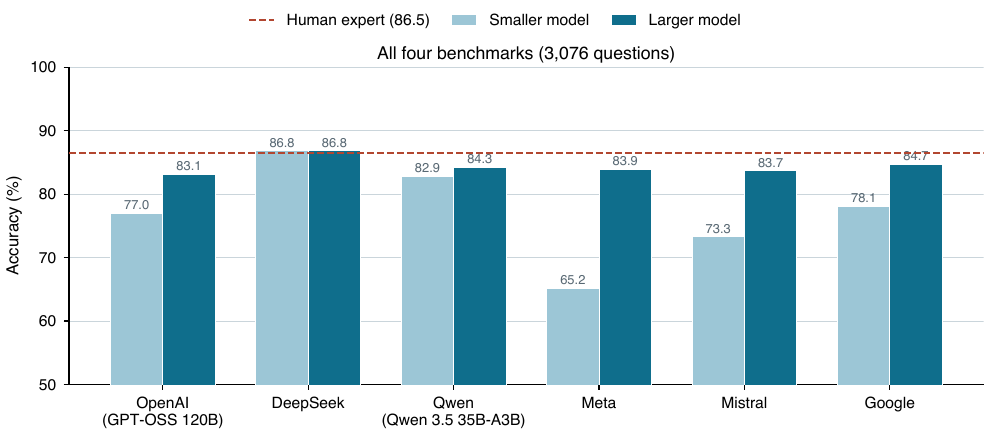}
\caption{Figure~\ref{fig:main1} with GPT-OSS 120B and Qwen 3.5 35B as the larger models of OpenAI and Qwen, in place of GPT-5.6 Luna and Qwen 3.7 Plus. Multiple-choice accuracy on all 3,076 questions of the four benchmarks, with each provider's smaller and larger model side by side. The dashed line is the human-expert reference of 86.5\%.}
\label{fig:supp-ow-size}
\end{figure}

\begin{figure}[H]
\centering
\includegraphics[width=0.71\linewidth]{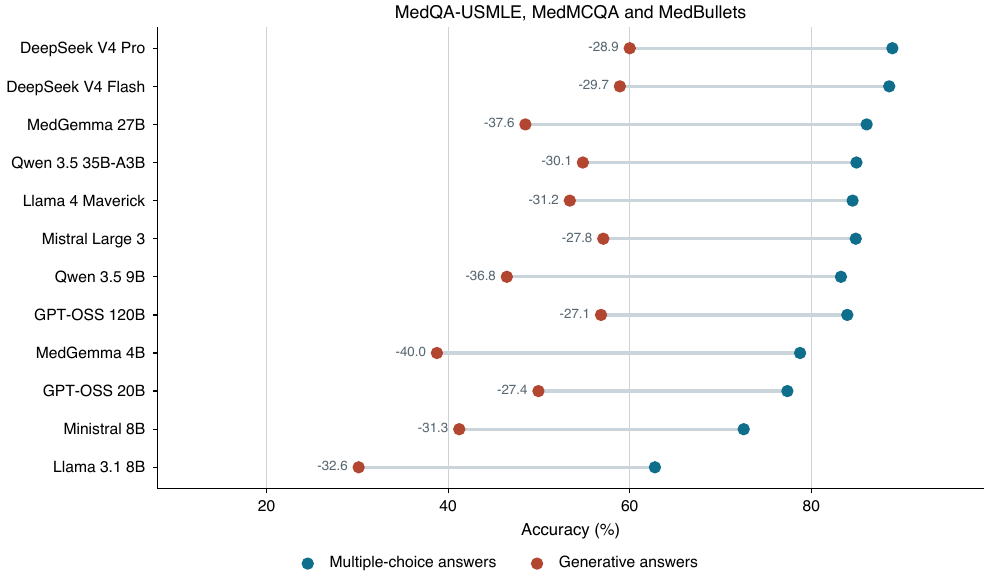}
\caption{Figure~\ref{fig:main2} with GPT-OSS 120B and Qwen 3.5 35B in place of GPT-5.6 Luna and Qwen 3.7 Plus. Accuracy on MedQA-USMLE, MedMCQA and MedBullets pooled (2,576 questions) in multiple choice (Setting 1) and the generative setting (Setting 2), one row per model; the segment gives the difference in points.}
\label{fig:supp-ow-open}
\end{figure}

For OpenAI and Qwen, whose larger models in the main text are closed and of undisclosed size, we also ran a third, open-weight model of reported size: GPT-OSS 120B (117B total and 5.1B active parameters) and Qwen 3.5 35B (35B total and 3B active). Figures~\ref{fig:supp-ow-size} and~\ref{fig:supp-ow-open} repeat Figures~\ref{fig:main1} and~\ref{fig:main2} with these two models in place of GPT-5.6 Luna and Qwen 3.7 Plus. The smaller-versus-larger trend holds and shifts only slightly. Across the six provider pairs, the larger model's average lead falls by 1.5 points in multiple choice (from 8.7 to 7.2), by 2.5 points without the answer choices (from 13.4 to 10.9) and by 2.1 points in reconstruction (from 7.2 to 5.1), and remains significant in all three (paired \mbox{$t$}-tests, \mbox{$p=.048$}, \mbox{$.018$} and \mbox{$.018$}). Qwen 3.5 35B has more parameters than Qwen 3.5 9B in total but fewer active per token, and it ran with its reasoning mode off, whereas Qwen 3.5 9B used it in Setting 1; with reasoning off for both, the Qwen lead in multiple choice is 4.3 points rather than 1.4.

\subsection*{S17. Expert evaluation of the LLM judge}
\phantomsection\label{sec:supp-expert}

\begin{figure}[H]
\centering
\includegraphics[width=0.72\linewidth]{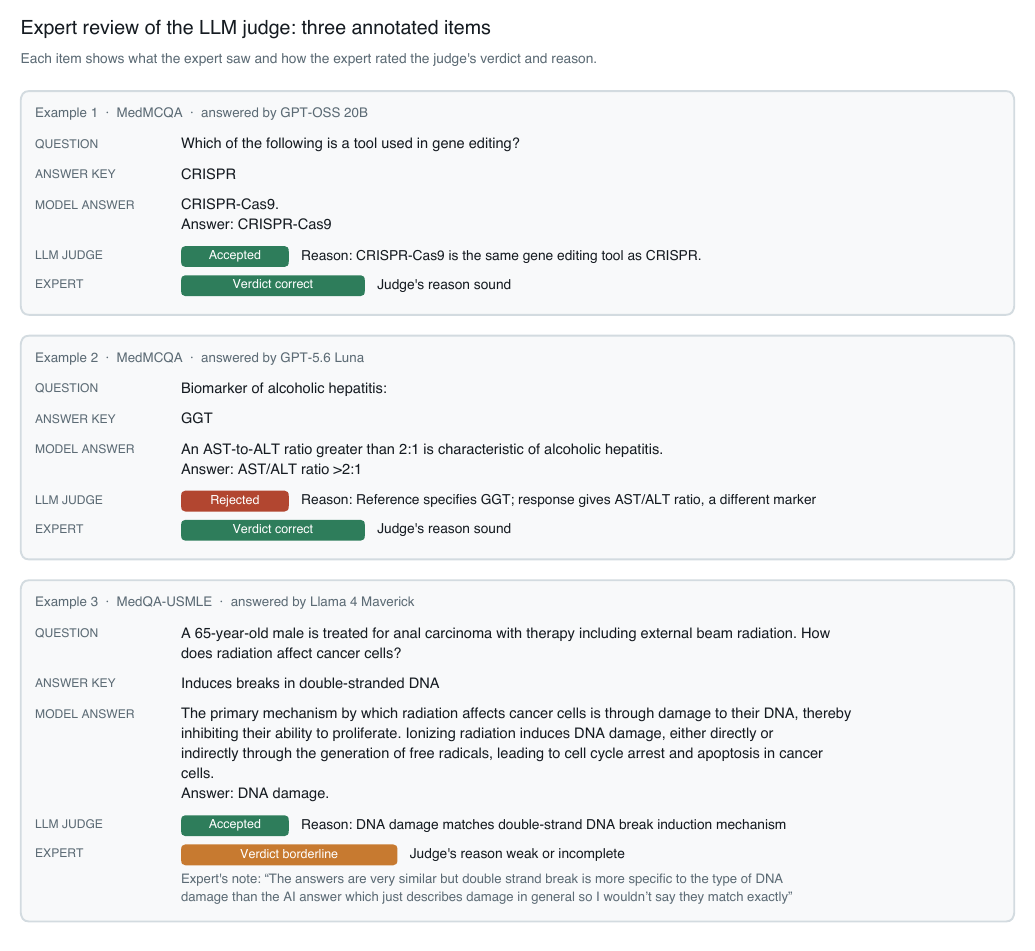}
\caption{Three items from the expert review of the LLM judge, as shown to the expert, with the expert's ratings. Examples 1 and 2 are an acceptance and a rejection the expert agreed with; Example 3 is the one acceptance the expert rated borderline. Model answers are shown in full.}
\label{fig:supp-annotation}
\end{figure}

Figure~\ref{fig:supp-annotation} shows three of the 50 items from the expert review in the Validation of the LLM judge section, laid out as on the expert's form. Each item gave the question, the answer key, the model's answer, and the judge's verdict with its reason. The expert then rated whether the verdict was correct (yes, no, or borderline) and whether the reason was sound (sound, weak or incomplete, or wrong or irrelevant), and could flag an answer key that looked wrong. Items were presented in random order, half accepted and half rejected by the judge. The first two examples are an acceptance and a rejection the expert agreed with. The third is the one item the expert did not fully agree with: an acceptance rated borderline, because the model's answer named DNA damage in general while the key specified breaks in double-stranded DNA.

\newpage

\bibliography{references}

\end{document}